\documentclass[journal]{IEEEtran}
\ifCLASSINFOpdf
\else
\fi

\usepackage{times}
\usepackage{epsfig}
\usepackage{graphicx}
\usepackage{amsmath}
\usepackage{amssymb}
\usepackage{multirow}
\usepackage{xcolor}
\usepackage{cite}

\usepackage{textcomp}
\usepackage{tabularx}

\usepackage[]{hyperref}

\begin{document}
%
\title{Rethinking and Designing A High-performing Automatic License Plate Recognition Approach}
%
%
%

\author{Yi~Wang*, \textit{Member, IEEE}, Zhen-Peng~Bian*, Yunhao~Zhou, and Lap-Pui~Chau, \textit{Fellow, IEEE}
\thanks{Yi~Wang, Yunhao~Zhou, and Lap-Pui~Chau are with School of  Electrical and Electronics Engineering, Nanyang Technological University, Singapore, 639798 (e-mail: wang1241@e.ntu.edu.sg, zh0022ao@e.ntu.edu.sg, elpchau@ntu.edu.sg).}
\thanks{Zhen-Peng Bian is with Singapore Telecommunications Limited, Singapore, 239732 (e-mail: zbian1@ntu.edu.sg).}
\thanks{Corresponding author: Lap-Pui Chau.}
\thanks{*Authors contributed equally.}
}

%
%

\markboth{IEEE TRANSACTIONS ON INTELLIGENT TRANSPORTATION SYSTEMS}%
{Shell \MakeLowercase{\textit{et al.}}: Bare Demo of IEEEtran.cls for IEEE Journals}
%



\maketitle

\begin{abstract}
In this paper, we propose a real-time and accurate automatic license plate recognition (ALPR) approach. Our study illustrates the outstanding design of ALPR with four insights: (1) the resampling-based cascaded framework is beneficial to both speed and accuracy; (2) the highly efficient license plate recognition should abundant additional character segmentation and recurrent neural network (RNN), but adopt a plain convolutional neural network (CNN); (3) in the case of CNN, taking advantage of vertex information on license plates improves the recognition performance; and (4) the weight-sharing character classifier addresses the lack of training images in small-scale datasets. Based on these insights, we propose a novel ALPR approach, termed VSNet. Specifically, VSNet includes two CNNs, i.e., VertexNet for license plate detection and SCR-Net for license plate recognition, integrated in a resampling-based cascaded manner. In VertexNet, we propose an efficient integration block to extract the spatial features of license plates. With vertex supervisory information, we propose a vertex-estimation branch in VertexNet such that license plates can be rectified as the input images of SCR-Net. In SCR-Net, we introduce a horizontal encoding technique for left-to-right feature extraction and propose a weight-sharing classifier for character recognition. Experimental results show that the proposed VSNet outperforms state-of-the-art methods by more than 50\% relative improvement on error rate, achieving $>$ 99\% recognition accuracy on CCPD and AOLP datasets with 149 FPS inference speed. Moreover, our method illustrates an outstanding generalization capability when evaluated on the unseen PKUData and CLPD datasets.
\end{abstract}

\begin{IEEEkeywords}
Convolutional neural network, license plate detection, character recognition, image classification, real-time system.
\end{IEEEkeywords}

%
\IEEEpeerreviewmaketitle

\section{Introduction}
\begin{figure} [t]
\centering 
\includegraphics[width=8.5cm]{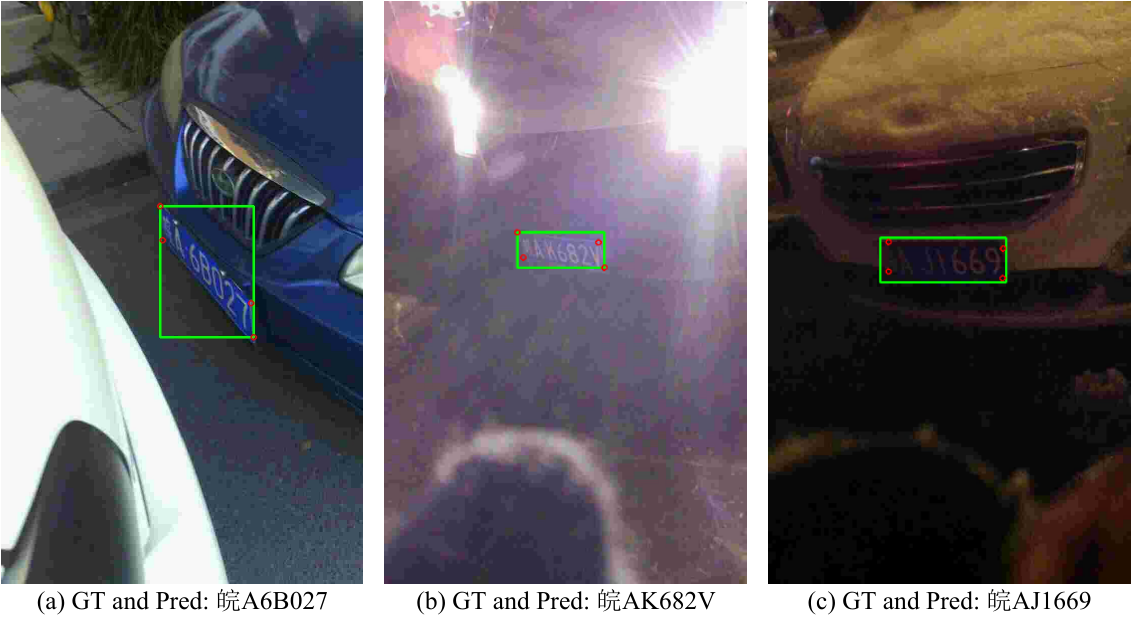}\\ 
\caption{Images captured in unconstrained scenarios, e.g., (a) oblique view, (b) uneven illumination, and (c) snow in the nighttime, labelled by ground-truth (GT) bounding boxes, four vertices, and characters. Our ALPR approach can handle these cases and generate correct predictions (Pred).}\label{Fig1} 
\end{figure}

\IEEEPARstart{A}{utomatic} license plate recognition (ALPR) aims at exploiting image processing and pattern recognition techniques to extract and recognize the characters on license plates (LPs) from vehicle images or videos. ALPR approaches are crucial for intelligent transport system (ITS) in a wide variety of applications, such as traffic surveillance and control, parking management, and toll station management \cite{anagnostopoulos2006license}. However, the ALPR methods often suffer from images captured in unconstrained scenarios, such as oblique views, uneven illumination, and different weather conditions, as shown in Fig. \ref{Fig1}. 

In recent years, many researchers have been paying attention to deep neural networks (DNNs) because of their powerful feature representation. Instead of devising hand-crafted features, they used DNNs to extract high-level discriminative features through data-driven learning strategies. This enables a dramatic improvement on computer vision tasks, including object detection \cite{zhao2019object}, semantic segmentation \cite{chen2017deeplab}, character recognition \cite{DAN_aaai20,luo2019moran}. ALPR methods benefit from the success of these tasks. Current ALPR methods either adopt a unified framework \cite{li2018toward} where two typical components, i.e., license plate detection and character recognition, are performed simultaneously in an end-to-end manner, or design a cascaded framework \cite{montazzolli2018license} where each component is optimized for specific tasks and connected in a cascaded manner. Regardless of which framework is chosen, the speed/accuracy trade-off always accompanies ALPR's design process. How to design an effective and efficient ALPR system is still an open-ended question.

In unified frameworks, the region proposal network (RPN) \cite{ren2015faster} was commonly adopted. Li \textit{et al.} \cite{li2018toward} proposed a unified method, named TE2E, which uses a RPN for LP detection and recurrent neural networks (RNNs) for character recognition. This method is not efficient due to RNNs. Xu \textit{et al.} \cite{xu2018towards} proposed a CNN-based roadside parking network (RPnet) to improve efficiency. However, small-size images adversely affect recognition performance since the details of the characters are lost. Enlarging input size implies an increase of the inference time. Another issue is that task-specific techniques (e.g., LP rectification for input images) are unavailable in such unified networks. These drawbacks harm both detection and recognition performance.

On the other hand, the typical DNN-based cascaded frameworks employ an LP detection network followed by the character recognition network. Reference \cite{laroca2019efficient} and \cite{montazzolli2018license} utilize YOLO \cite{redmon2017yolo9000} networks to perform LP detection and use CR-NET \cite{silva2017real} for character segmentation and recognition. A semantic segmentation network was proposed to segment characters on LPs by Zhuang \textit{et al.} \cite{zhuang2018towards}, but it requires laborious annotations of each character. Other recent methods \cite{wang2019light,zhang2020robust,zou2020robust} adopt RNN-based structures for character recognition. However, RNN hinders a compact and efficient design. Additional character segmentation or RNNs decreases the inference speed compared with CNNs.

\begin{figure} [t]
\centering 
\includegraphics[width=8.5cm]{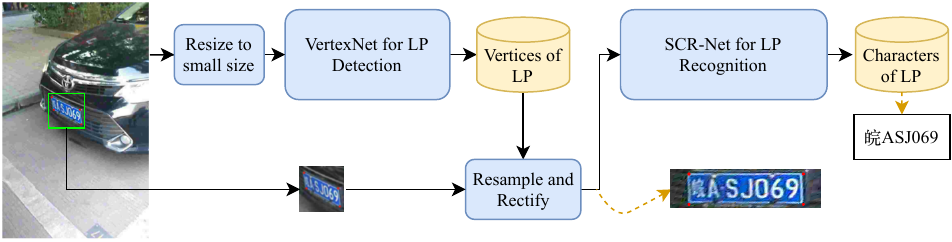}\\ 
\caption{Framework of the proposed VSNet. An input image is resized to a small resolution, i.e., 256$\times$256, for fast inference in VertexNet. Then, the LP patch is resampled from the finest input image and rectified to high resolution according to the predicted vertices by VertexNet. Finally, SCR-Net recognizes all characters in the LP.}\label{Fig2} 
\end{figure}

In view of these issues, we propose a high-performing ALPR approach, termed VSNet, with four insights. 
\begin{enumerate}
\item We propose a resampling-based cascaded manner to combine the detection and recognition components, i.e., VertexNet for LP detection and Squeeze Character Recognition Network (SCR-Net) for LP recognition, as shown in Fig. \ref{Fig2}. VertexNet is designed with small-resolution input (even losing character details), achieving high inference speed and decreasing memory usage. This would not affect recognition performance in SCR-Net since the LP are resampled from the finest input image and rectified to high resolution according to the predicted vertices by VertexNet.
\item Our network architecture adopts well-designed CNNs without adopting character segmentation or RNNs. One forward pass of the network can obtain detection and recognition results, which is more suitable for the real-time system with limited resources. The first two insights ensure fast inference speed.
\item We separately implement VertexNet and SCR-Net with novel vertex-based architecture designs. In VertexNet, we propose an integration block (IB) to effectively extract the spatial features of LPs. With vertex supervisory information, we propose a vertex-estimation branch for predicting the geometric shapes of LPs, which are further utilized for LP rectification in SCR-Net. Moreover, the vertices can be used to augment training images. In SCR-Net, we introduce a horizontal encoding technique for left-to-right feature extraction and propose a weight-sharing classifier for character recognition.
\item We propose a weight-sharing classifier to balance the training samples and address the failure of character recognition on small-scale datasets. Our weight-sharing classifier only account for 4\% parameters of the fully-connected classifier. The last two insights guarantee high accuracy.
\end{enumerate}

Table \ref{table1} illustrates the comparisons of the characteristics between our VSNet and state-of-the-art methods. Compared with unified frameworks \cite{li2018toward,xu2018towards}, and RNN-based methods \cite{wang2019light,zhang2020robust,zou2020robust}, the proposed VSNet employs the cascaded framework with the CNN structure only and adopts the vertex estimation and LP rectification, obtaining the best inference speed and accuracy. Experiments show that VertexNet improves the detection precision to 99.1\% on the CCPD \cite{xu2018towards} dataset, surpassing the state-of-the-art method \cite{wang2019light} by 60\% relative improvement on error rate. SCR-Net increases the recognition accuracy to 99.5\% and 99.7\% on the CCPD and AOLP \cite{hsu2012application} datasets, respectively, achieving more than 50\% relative improvement on error rate compared with the recent work \cite{zhang2020robust}. The cross-dataset tests on PKUData \cite{yuan2016robust} and CLPD \cite{zhang2020robust} illustrate that our method has an outstanding generalization capability, showing more than 90\% accuracy even without training on these datasets. Moreover, we validate that the vertex estimates help CNN-based character recognizer such that it avoids adopting additional character segmentation networks or RNNs. Therefore, our method has the fastest inference speed among the existing ALPR methods, i.e., the fast version of our approach can run at 149 FPS on a single NVIDIA GTX1080Ti GPU. 

\begin{table}[t]
\centering \caption{Comparisons of the characteristics between state-of-the-art methods and VSNet. Performance is evaluated on CCPD \cite{xu2018towards}.}
\label{table1}
\begin{tabular}{lcccccc}
\hline
Characteristic    &  \cite{li2018toward}      & \cite{xu2018towards}      &  \cite{wang2019light} & \cite{zhang2020robust} & \cite{zou2020robust}   & VSNet       \\ \hline
Unified framework    & \checkmark & \checkmark          &            &   &    &       \\ \hline
Cascaded framework    &            &           & \checkmark  & \checkmark & \checkmark  & \checkmark \\ \hline
RNN structure & \checkmark           &           & \checkmark & \checkmark& \checkmark &  \\ \hline
CNN structure only     &   & \checkmark    &  &    &  & \checkmark      \\ \hline
Vertex estimation &            &                 & \checkmark  & & & \checkmark \\ \hline
LP rectification     &            &            &     \checkmark       &   & & \checkmark \\ \hline
Inference speed rank   &   5  & 2  & 3 & 4 & - & \textbf{1} \\ \hline
Accuracy rank   &    6  & 5  & 3 & 2 & 4 & \textbf{1} \\ \hline
\end{tabular}
\end{table}

The rest of this paper is organized as follows. We first introduce the related works in Section \ref{Related}. Then, the proposed method is presented in detail in Section \ref{Method}. Experimental results and ablation studies are discussed in Section \ref{Experiment}. Finally, we conclude this paper in Section \ref{Conclusion}.

\section{Related Work} \label{Related}

The scope of automatic license plate recognition (ALPR) in this paper is the task of automatically localizing and recognizing license plates (LPs) in images. The typical ALPR is comprised of two components, i.e., LP detection and LP recognition. In this section, we first review LP detection and LP recognition methods, and then we present the unified framework and cascaded framework in ALPR systems.

\subsection{License Plate Detection}

As a basic component, LP detectors play an important role in localizing LPs, by which the bounding boxes are produced to encompass each LP. In traditional techniques, LP detection is based on features extraction manners \cite{du2012automatic}, such as edge, color, texture, character features, or local binary patterns (LBPs \cite{al2018ensemble}). However, these hand-crafted features are sensitive to the complex background, e.g., objects with similar shape, color, texture, or characters. Recently, CNNs have dominated object detection \cite{zhao2019object} for years. Given sufficient training data, CNNs have powerful feature representation and high performance compared with traditional hand-crafted feature extraction methods. Here we focus on CNN-based detection methods. Typically, LP detection can be partitioned into two groups: two-stage LP detection and one-stage LP detection. 

\subsubsection{Two-Stage LP Detection}
Fast R-CNN \cite{ren2015faster} and Faster R-CNN \cite{girshick2015fast} are baseline works of two-stage detection networks, which generate high-recall region proposals at the first stage and refine them at the second stage. TE2E \cite{li2018toward} employed RPN \cite{ren2015faster} to generate high-quality bounding box proposals, and used the RoI pooling layer to obtain region features for LP detection. Dong \textit{et al.} \cite{dong2017cnn} modified the RPN structure by the inception block 3a and 3b of GoogLeNet v2 \cite{szegedy2016rethinking}, and used R-CNN to regress four corner points. However, the two-stage detection is less efficient than the one-stage detection. Most LP detection methods adopted the one-stage detection framework.

\subsubsection{One-Stage LP Detection}
One-stage detection networks, such as SSD \cite{liu2016ssd}, YOLO \cite{redmon2016you}, and YOLOv2 \cite{redmon2017yolo9000}, predict class probabilities and bounding box offsets in a single forward pass, satisfying the trade-off of accuracy and speed in the ALPR system. YOLO networks are prevalent in LP detection due to their high efficiency. Silva and Jung \cite{silva2017real} are pioneers who introduced two cascaded YOLO networks to detect LPs. Owing to the small size of LPs, the vehicle's frontal view is detected first, and then the LPs are detected. This method shows high precision and recall rates on the Brazilian SSIG database \cite{gonccalves2016benchmark}. To detect rotated LPs, Xie \textit{et al.} \cite{xie2018new} proposed an MD-YOLO framework for multi-directional LP detection. A prepositive CNN model is to determine the attention region of an LP. The MD-YOLO takes a cropped attention region as the input and finally produces the accurate rotational box, which achieves high detection accuracy on the AOLP \cite{hsu2012application} dataset. However, MD-YOLO only considers on-plane rotations. To handle complex deformations of LPs, Silva and Jung \cite{montazzolli2018license} adopted the YOLOv2 network for vehicle detection and proposed a Warped Planar Object Detection Network (WPOD-NET) to detect LPs and regress their transformation parameters, which shows rectified LPs can be better recognized. Both MD-YOLO and WPOD-NET are 2D transformation of LPs. Other works, like \cite{laroca2019efficient,hsu2017robust,laroca2018robust}, also adopted the modified YOLO, YOLOv2 networks.

As another type of one-stage detection method, SSD \cite{liu2016ssd} is an efficient approach where multi-scale feature maps are exploited to adapt to multi-scale objects in an image. Xu \textit{et al.} \cite{xu2018towards} introduced multi-scale feature maps for box regression and character recognition. The multi-scale features can help to detect objects on different scales. Based on SSD, Chen \textit{et al.} \cite{chen2019simultaneous} proposed a multi-branch attention neural network to detect vehicles and LPs simultaneously. Based on the Cascaded CNN \cite{qin2016joint}, Wang \textit{et al.} \cite{wang2019light} proposed a multi-task CNN (MTCNN) to detect LPs, i.e., plate classification, bounding box regression, plate landmark localization, and plate color recognition. This method enhances the detection precision on the CCPD dataset. 

\begin{figure*} [t]
\centering 
\includegraphics[width=17.5cm]{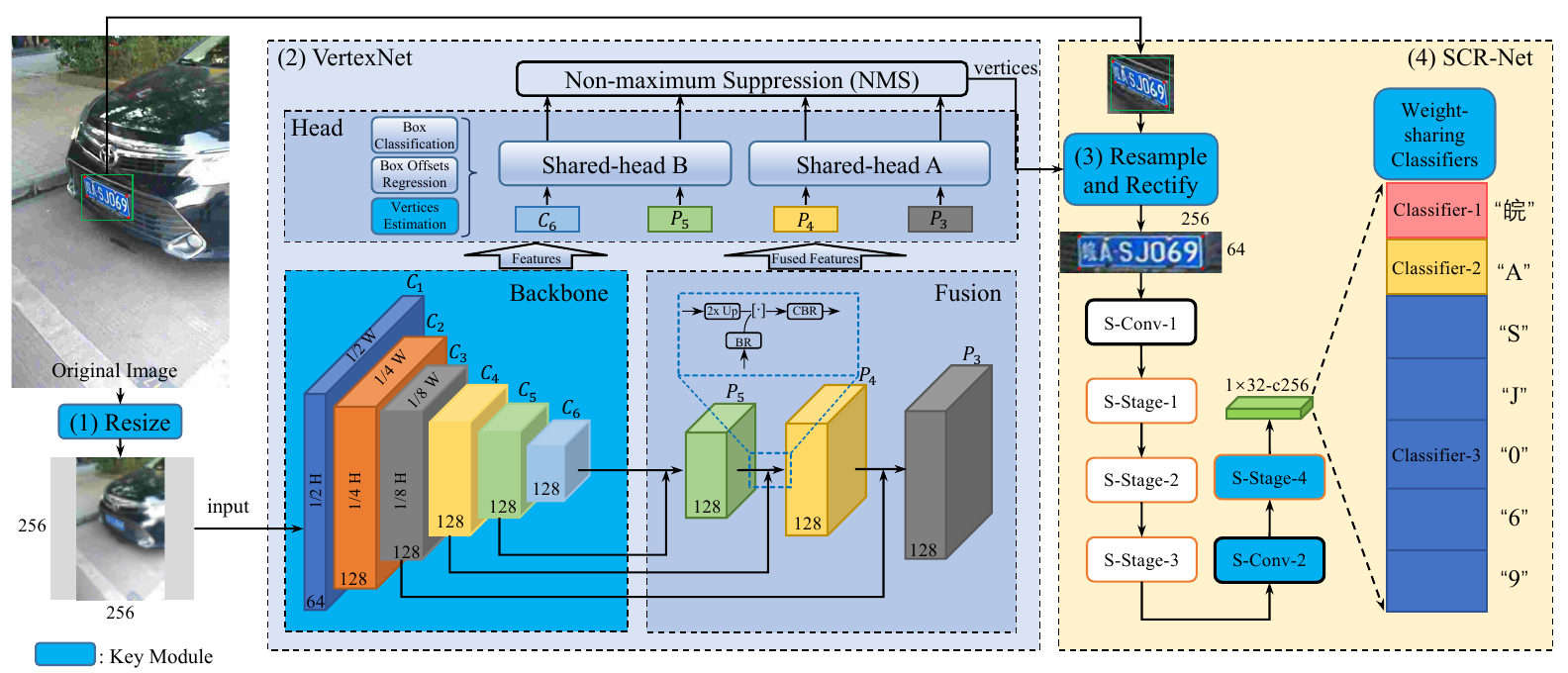}\\ 
\caption{Architecture of VSNet. (1)-(4) are executed sequentially. VertexNet consists of the backbone, fusion, and head networks, which predicts the bounding boxes and vertices of LPs. In the fusion structure, ``2$\times$Up'' is the bilinear upsampling with a factor of 2. ``BR'' is the batch normalization followed by the ReLU. ``CBR'' is the 1x1-s1 convolution followed by the ``BR''. $[\cdot]$ is the concatenation operation. SCR-Net resamples and rectifies LP images from the finest input images based on the predicted vertices (red points in the LP) and recognizes all characters. Zooming in the figure for better viewing.}\label{Fig3} 
\end{figure*}

\subsection{License Plate Recognition}
The ultimate goal of the ALPR system is to recognize all letters and numbers on a given LP. There are two typical categories of LP recognition: segmentation-based recognition and segmentation-free recognition. 

\subsubsection{Segmentation-based Recognition}
Segmentation-based recognition methods first segment an LP into character regions, and then the regions are recognized one-by-one via optical character recognition (OCR) techniques \cite{li2018toward,montazzolli2018license}. In \cite{hsu2012application}, the maximally stable extreme region (MSER)  was proposed to segment characters, and the LBP features are extracted and classified by a linear discriminant analysis (LDA) classifier. Gou \textit{et al.} \cite{gou2015vehicle} proposed an AdaBoost classifier with decision trees to select the extremal regions (ERs) as character regions and proposed a hybrid discriminative restricted Boltzmann machines (HDRBM) to recognize the characters. However, these methods suffer from segmentation errors, which hinders their applications in unconstrained environments. A semantic segmentation network was proposed to segment characters by Zhuang \textit{et al.} \cite{zhuang2018towards}. Despite improving the recognition accuracy, the method is achieved by laborious data annotations, i.e., annotations for each character on natural images and 60,000 synthetic images. In this paper, we focus on segmentation-free recognition methods.

\subsubsection{Segmentation-free Recognition}
This type of approach directly performs LP recognition without using character segmentation. LP recognition can be referred to as sequence recognition, and RNN-based structures were widely adopted. In TE2E \cite{li2018toward}, Li \textit{et al.} proposed a bidirectional RNNs (BRNNs) with connectionist temporal classification (CTC) loss \cite{graves2008novel} to process LP features produced by RPN. Similarly, Wang \textit{et al.} \cite{wang2019light} proposed a convolutional RNN (CRNN) followed by the CTC for character recognition. Combining with the abovementioned MTCNN detection, they proposed a multi-task license plate detection and recognition (MTLPR) model. Zou \textit{et al.} \cite{zou2020robust} introduced a Bi-LSTM to locate the character positions implicitly without segmenting each character, and they used a 1D-Attention to extract features of the character regions. Zhang \textit{et al.} \cite{zhang2020robust} proposed a tailored Xception network for feature extraction and a 2D-attention based RNN for character recognition, named Attentional Net, which obtains state-of-the-art performance on both constrained and unconstrained environments. Generally, RNN-based structures are less efficient than CNN-based methods when they have the same number of parameters. In RPnet \cite{xu2018towards}, Xu \textit{et al.} employed a set of CNN-based classifiers to recognize characters in predefined positions. Although the RPnet improves the inference speed, its performance is inferior to the RNN-based methods, such as MTLPR and Attentional Net. Unlike previous works, our approach adopts CNN only and has high recognition performance via a novel architecture design.

\subsection{ALPR Framework}
As a systematical design, the ALPR framework is a combination manner of functional components, such as LP detection, character segmentation, and character recognition. The approach of combining these components significantly affects the performance of ALPR systems. We roughly classify the ALPR framework into two categories, the unified framework and the cascade framework. The unified frameworks \cite{li2018toward,xu2018towards} predict LPs' location and recognize characters simultaneously. It is an end-to-end trainable structure. However, these frameworks are limited to architecture design. Recognizing small characters needs to increase the input size, thereby decreasing the inference speed. Another issue is those task-specific techniques cannot fit both detection and recognition well. For example, data augmentation needs to be performed in different forms for detection and recognition tasks. 

The typical cascaded framework is a stack of LP detection, character segmentation, and character recognition components, each of which is optimized by task-specific techniques. Since the character segmentation is prone to complex environments, such as uneven lighting, it is omitted in most DNN-based approaches. With the advances of DNNs, plate detection directly followed by recognition has emerged accordingly, such as \cite{silva2017real,wang2019light,zhang2020robust}. The cascaded frameworks are subject to off-the-shelf detection architectures, like YOLO and SSD, which is not an optimal ALPR scheme. Our method is built on the resampling-based cascaded framework and optimizes several components, as described in Section \ref{Method}.

\section{Proposed Method}\label{Method} 

The proposed VSNet is composed of two cascaded CNNs, i.e., VertexNet for LP detection and SCR-Net for LP recognition. Fig. \ref{Fig3} shows the architecture of VSNet. We decouple the design of VSNet into detection and recognition, improving overall performance. VertexNet is a compact yet high-performing one-stage detector implemented by small input size, narrow channel of high-level layers, and vertex estimation. SCR-Net is a robust and efficient classifier. We propose LP rectification to resample and normalize various LP detections. The architecture employs a horizontal encoding technique for left-to-right feature extraction and a weight-sharing classifier for character recognition. This section will describe the proposed VertexNet and SCR-Net in detail. 

\subsection{VertexNet for LP Detection}
The inference speed of a detector is considered as a bottleneck of VSNet. The critical insight is to design the detector with a small input size and compact architecture. However, the small input size leads to extremely small LPs, which poses a threat to state-of-the-art detectors. Some previous methods \cite{laroca2019efficient} applied two cascaded YOLO networks with the input size of 416$\times$416 to detect vehicles and LPs sequentially, but two-step detection increases the inference time and may accumulate errors. In VertexNet, we instead introduce a single detector with the small size (256$\times$256). To deal with small LPs, we propose a novel architecture.

\subsubsection{Architecture of VertexNet} 
The proposed VertexNet contains three modules, i.e., backbone, fusion, and head networks, as shown in Fig. \ref{Fig3}. Specifically, instead of using off-the-shelf architectures (YOLO or SSD) as a backbone, we propose an integration block (IB) by leveraging the residual block \cite{he2016deep} and enhancing the Squeeze-and-Excitation (SE) attention module \cite{hu2018squeeze}. In the head network, in addition to box-offsets estimation, we propose a vertex-estimation branch of LPs, which improves the detection performance and benefits to character recognition in SCR-Net.

\textbf{Backbone network.} The backbone is split into six stages (Stage-1 to 6), and the output size is half of the input size in each stage. We define the output feature maps as \(C_{i}, i \in \{1,2,...,6\}\). The input images are first fed to Stage-1, which employs a 7$\times$7-s2 Conv-BN-ReLU layer, i.e., convolutional layer with the kernel of 7$\times$7 and the stride of 2 followed by the batch normalization and rectified linear unit. Then, the output features are fed to the last five stages successively, each of which is composed of two IBs. The downsampling of feature maps is achieved by a 2$\times$2-s2 max-pooling layer before the first IB in Stage-2, while the downsampling of feature maps is achieved by a 3$\times$3-s2 convolutional layer inside the first IB in the remaining stages.

\begin{figure} [t]
\centering 
\includegraphics[width=8.5cm]{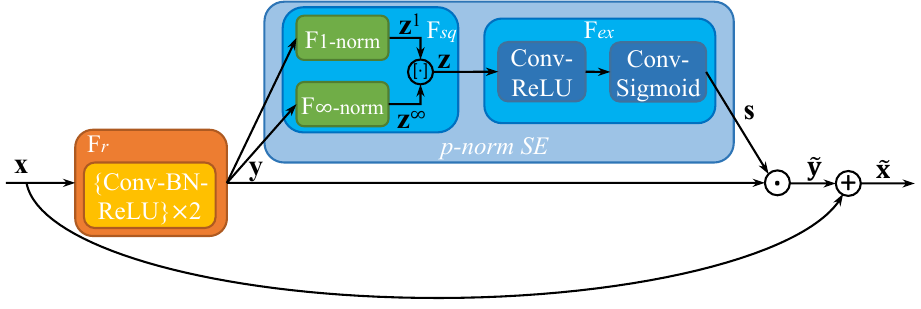}\\ 
\caption{Implementation of the integration block. $\mathrm{F}_{r}$, $\mathrm{F}_{sq}$, and $\mathrm{F}_{ex}$ are the residual branch, squeeze step, and excitation step, respectively. $[\cdot]$,  $\odot $, and $\oplus$ denote the concatenation operation, element-wise product, and element-wise addition, respectively. \(\{\cdot\}\times x\) denotes performing the \(\{\cdot\}\) structure by \(x\) times. The convolutional layers have the kernel size of 3$\times$3 in $\mathrm{F}_{r}$ and 1$\times$1 in $\mathrm{F}_{ex}$.}\label{Fig4} 
\end{figure}

The IB is an effective building block in VertexNet, as shown in Fig. \ref{Fig4}. Since LPs present a small scale, the multiple channels in top layers of ResNet \cite{he2016deep} that handles large objects are unnecessary for small LP detection. We design the IB with fixed-channel residual block \cite{he2016deep}. The number of channels in each IB is set to 128. This naive version of IB contains fewer parameters, and its accuracy is comparable to ResNet-50 (see Section \ref{abl-vertexnet}). Moreover, we enhance the SE attention module \cite{hu2018squeeze} by introducing the $p$-norm in the \textit{squeeze} step, named $p$-norm SE. The \textit{squeeze} step aggregates the feature maps across spatial dimensions to generate channel descriptor. In the original SE module, only the global average pooling is adopted. In the $p$-norm SE module, however, we calculate two channel-wise $p$-norms, i.e., $1$-norm and $\infty$-norm, and then fuse those features. The $1$-norm feature encodes the global average information, while the $\infty$-norm feature extracts the significant information. The $p$-norm SE enhances the channel-wise statistics compared with the original SE. With the $p$-norm SE, the IB is superior to ResNet-50 in terms of accuracy and model size (see Section \ref{abl-vertexnet}).

Formally, let the input and output features of the residual branch ($\mathrm{F}_{r}$) be $\mathbf{x}$ and $\mathbf{y}$, respectively, as shown in Fig. \ref{Fig4}. We have $\mathbf{y}\in \mathbb{R}^{C\times H\times W}$ and $\mathbf{y}=[y_{1}, y_{2}, ..., y_{C}], y_{c} \in \mathbb{R}^{H\times W}$ where $y_{c}$ is spatial features of the $c$-th channel ($C$=128 for all experiments). We first feed the features to the \textit{squeeze} step ($\mathrm{F}_{sq}$) of the $p$-norm SE and obtained two channel descriptors $\mathbf{z}^{k} \in \mathbb{R}^{C\times 1\times 1}=[z^{k}_{1}, z^{k}_{2}, ..., z^{k}_{C}], k=\{1,\infty\}$, which are calculated by Eq. (\ref{eqn1}) and Eq. (\ref{eqn2}), respectively:
\begin{equation}
\label{eqn1}
z_{c}^{1}=\mathrm{F}_{\mathrm{1-norm}}(y_{c})=\sum_{i=1}^{H}\sum_{j=1}^{W}y_{c}(i,j),
\end{equation} and
\begin{equation}
\label{eqn2}
z_{c}^{\infty}=\mathrm{F}_{\mathrm{\infty-norm}}(y_{c})=\max_{i,j}y_{c}(i,j).
\end{equation}
In practice, we adopt the global average-pooling layer and global max-pooling layer to implement the $1$-norm and $\infty$-norm squeeze operation. Then, the channel descriptor $\mathbf{z} \in \mathbb{R}^{2C \times 1\times 1}$ is obtained by concatenating $1$-norm and $\infty$-norm features, i.e., $\mathbf{z} = [\mathbf{z}^{1}, \mathbf{z}^{\infty}]$. Next, we perform the \textit{excitation} step ($\mathrm{F}_{ex}$), like the one in \cite{hu2018squeeze}, to capture channel-wise dependencies. Specifically, a 1$\times$1 Conv-ReLU followed by a 1$\times$1 Conv-Sigmoid is applied to produce the channel descriptor:
\begin{equation}
\label{eqn3}
\mathbf{s}=\mathrm{F}_{ex}(\mathbf{z}, \mathbf{W})=\sigma (\mathbf{W}_{2}\times\delta (\mathbf{W}_{1}\times \mathbf{z})),
\end{equation}
where $\mathbf{W}_{1} \in \mathbb{R}^{\frac{C}{r}\times 2C}$ and $\mathbf{W}_{2}\in \mathbb{R}^{C \times \frac{C}{r}}$ are the weights of the first and second convolutional layers, $\delta$ and $\sigma$ denote the ReLU and sigmoid activation, respectively, and $\mathbf{s}\in \mathbb{R}^{C \times 1\times 1}$ is the scalar to control the channel attention of the features $\mathbf{y}$. $r$ is the reduction ratio and is set to $4$ for all experiments. The output of the $p$-norm SE is calculated by scaling $\mathbf{y}$ with $\mathbf{s}$:
\begin{equation}
\label{eqn4}
\mathbf{\tilde{y}}={\mathbf{s}}' \odot \mathbf{y},
\end{equation}
where $\mathbf{\tilde{y}}\in \mathbb{R}^{C\times H\times W}$ is the scaled features, $\odot$ denotes the element-wise product, and ${\mathbf{s}}'$ repeats the values of $\mathbf{s}$ in $H$ and $W$ dimensions, i.e., $\mathbb{R}^{C \times 1\times 1}\rightarrow\mathbb{R}^{C \times H\times W}$. Finally, the output features of IB is defined by:
\begin{equation}
\label{eqn5}
\mathbf{\tilde{x}}=\mathbf{\tilde{y}} \oplus  \mathbf{x},
\end{equation}
where $\oplus$ is the element-wise addition between two tensors.

\textbf{Fusion network.} The feature pyramid network (FPN) \cite{lin2017feature} is modified to fuse the features of the backbone network. We employ a top-down route to fuse multi-scale features. As shown in the Fusion part of Fig \ref{Fig3}, the basic structure fuses the output features from two consecutive stages. To match the feature sizes, the bilinear upsampling is first used for upsizing the lower-resolution features by a factor of 2. The higher-resolution features are passed through BN-ReLU. Then, the two feature maps are concatenated and produced by a 1$\times$1-s1 Conv-BN-ReLU. With the concatenation and Conv-BN-ReLU, the structure becomes a weighted fusion manner, and the weights are learnable. We fuse the features from Stage-6 to Stage-3, producing the fused feature maps \(P_{i}, i \in \{5,4,3\}\). The fusion network has good ability in multi-scale feature representation.

\textbf{Head network.} In the previous works \cite{li2018toward,xu2018towards}, the head network performs LP classification and box offsets prediction. In addition, we add a vertex-estimation branch to predict four 2D vertices of LPs, as shown in the Head part of Fig \ref{Fig3}. Specifically, the head network takes three fused feature maps (\(P_{i}, i \in \{3,4,5\}\)) and one top-level feature maps (\(C_{6}\)) as the inputs and produces 4-scale predictions using two convolutional layers. Unlike SSD \cite{liu2016ssd}, the layers in our head network share the parameters across different-scale features, i.e., two convolutional layers process four feature maps. Large-scale feature maps \(P_{3}\) and \(P_{4}\) are processed by a 3x3-s1 convolutional filter (Shared-head A), while small-scale feature maps \(P_{5}\) and \(C_{6}\) are processed by another 3x3-s1 convolutional filter (Shared-head B). Sharing parameters not only allow the filter to fit different-scale objects but also decrease the network's parameters. 

The channel number of output predictions are based on the number of the default boxes (or anchor boxes). Following \cite{lin2017feature}, we use 9 anchor boxes for each location in the output feature maps. Since we predict 2 class scores (LP or background), 4 box offsets, 8 vertex values (4\(\times\)2D vertex), the channel number is equal to \((2+4+8)\times 9=126\). 

\textbf{VertexNet-Fast.} To achieve faster inference, we further simplify VertexNet with shallow architecture. In particular, VertexNet-Fast adopts the naive IB in the backbone network, i.e., six naive IB-based stages without the $p$-norm SE module. Although the SE module brings a few parameter increases, it doubles the network depth and increases the inference time.

\subsubsection{Training of VertexNet} 
We define the anchor boxes whose Jaccard index (or intersection over union, IoU) with the ground-truth bounding boxes larger than 0.25 as positive anchors. Other anchor boxes are defined as negative anchors. We choose such a small IoU threshold because it guarantees that most LPs can match the anchor boxes. Following SSD, we use hard negative mining to reduce the significant number of negative anchors. We choose negative anchors by the highest losses, and their number is 3 times larger than the positive ones. The overall loss consists of three aspects, i.e., classification loss for LP classification ($L_{cls}$), localization loss for box offset regression ($L_{box}$) , and vertex estimation loss for vertex regression ($L_{vert}$). Formally, we minimise the overall loss for any anchor box $j$:
\begin{equation}
L=\lambda_{1} L_{cls}(p_{j},\widehat{p}_{j})+\lambda_{2}p_{j}L_{box}(t_{j},\widehat{t}_{j}) + \lambda_{3}p_{j}L_{vert}(l_{j},\widehat{l}_{j})
\end{equation}
$L_{cls}$ is implemented by Softmax layer followed by the cross-entropy loss. $p_{j}$ is the ground-truth label of anchor $j$ with 1 standing for LP and 0 for background, and $\widehat{p}_{j}$ is the corresponding predicted probability. $L_{box}$ and $L_{vert}$ are the Smooth-L1 loss \cite{girshick2015fast}. $t_{j}=\{t_{x}, t_{y}, t_{w}, t_{h}\}$ and $l_{j}=\{l_{x_{1}}, l_{y_{1}}, ... , l_{x_{4}}, l_{y_{4}}\}$ are the ground-truth box coordinates and 2D vertices associated with the positive anchor, respectively. $\widehat{t}_{j}$ and $\widehat{l}_{j}$ are the predicted ones. $\lambda_{1}$, $\lambda_{2}$, and $\lambda_{3}$ are the hyper-parameters to balance these losses, and we empirically set them to 4, 0.8, and 0.1, respectively. 

\textbf{Remark.} Although our VertexNet predicts both bounding boxes and vertices of LPs, it can work smoothly with only vertex annotations since the boxes can be generated according to the vertices. The vertices imply the geometric information of LPs captured from various views, and the time cost of vertex annotations is equivalent to that of bounding boxes. We will validate that the vertex annotations are more suitable for LP recognition in Section \ref{abl-scrnet}.

\subsection{SCR-Net for LP Recognition}
Different from YOLO-based detection methods \cite{laroca2019efficient,silva2017real} or semantic segmentation method \cite{zhuang2018towards} that requires character-level annotations, we consider LP recognition as a classification problem for the predefined characters. First, the LP is resampled from the finest input image and rectified to high resolution (64$\times$256) according to its vertices. This step implicitly lays out and normalizes the locations of characters. Then, the rectified LP image is fed to SCR-Net. SCR-Net predicts the characters via a forward pass. In this section, we describe LP rectification, architecture, and training method.

\subsubsection{License plate rectification}
We resample and rectify LPs according to the predicted vertices from VertexNet. To be specific, the rectification process is achieved by perspective transformation which generates a bird's-eye view of the LPs regardless of the original camera angles. Suppose there is a vertex in an image with coordinates \([u, v, 1]\). Thus we can project the original vertex into the target 3D space by:
\begin{equation}
\label{eqnM}
 [\rho^{'}, \eta^{'}, \xi^{'}] = [u, v, 1]\times M= [u, v, 1]\begin{bmatrix}
a_{11} & a_{12} & a_{13}\\ 
a_{21} & a_{22} & a_{23}\\ 
a_{31} & a_{32} & 1
\end{bmatrix},
\end{equation}
where \(M\) denotes the transformation matrix with eight unknown variables. Next, a squeeze operation is used in order to express the target vertex in a plane which is perpendicular to camera lens. Specifically, the target vertex coordinates \( [\rho, \eta]\) are \( \rho = \rho^{'} / \xi^{'}\) and \(\eta = \eta^{'} / \xi^{'}\).

The projection between four vertices in the original plane and the target plane constructs eight equations through Eq. (\ref{eqnM}), which can be solved with linear algebra. Finally, we apply the same projection to all the pixels in the original plane. To ensure all characters are included in rectification, we enlarge the sampling width and height to $1.25  w_{l}$ and $1.25  h_{l}$, where \(w_{l}\) and \(h_{l}\) denote the width and height of an LP, respectively. Note that the LP rectification not only rectifies the view angle but also normalize the resampled LP to a predefined size, as shown in the step (3) of Fig. \ref{Fig3}.

\begin{table}[t]
\centering \caption{Descriptions of SCR-Net. ``Size'' and ``CH'' are the output size and channel of feature maps, respectively. Conv3: 3$\times$3 convolution. BN: batch normalization. Conv4: 4$\times$4-s2 convolution for downsampling the feature maps by a factor of 2. ReLU: rectified linear unit. Dy: dynamic regularization \cite{wang2020convolutional}. Conv8$\times$1: 8$\times$1-s1 convolution for a horizontal encoding. Conv1$\times$3: 1$\times$3-s1 convolution for 1$\times$32 features encoding. Res-block: residual block \cite{he2016deep} with 1$\times$3 convolution kernel. \(\{\cdot\}\times x\) denotes performing the \(\{\cdot\}\) structure by \(x\) times.}
\label{table2}
\resizebox{\columnwidth}{!}{%
\begin{tabularx}{9cm}{lccl}
\hline
Name    & Size   & Ch  & Descriptions                                                                                               \\ \hline
Input   & 64$\times$256 & 3   & Input image                                                                                                \\ \hline
S-Conv-1  & 32$\times$128 & 16  & Conv4-BN-ReLU                                                                                     \\ \hline
S-Stage-1 & 32$\times$128 & 93  &$\{$BN-Conv3-BN-ReLU-Conv3-BN-Dy$\}$\(\times\)4                                                                           \\ \hline
S-Stage-2 & 16$\times$64  & 176  &$\{$BN-Conv3-BN-ReLU-Conv3-BN-Dy$\}$\(\times\)4                                                                           \\ \hline
S-Stage-3 & 8$\times$32   & 256 &$\{$BN-Conv3-BN-ReLU-Conv3-BN-Dy$\}$\(\times\)4                                                                           \\ \hline
S-Conv-2  & 1$\times$32   & 256 &Conv3-BN-ReLU-Conv8$\times$1-BN-ReLU                                                                                    \\ \hline
S-Stage-4 & 1$\times$32   & 256 &$\{\{$Res-block$\}$\(\times\)2-Conv1$\times$3-BN-ReLU$\}$\(\times\)2                                                                                  \\ \hline
Classifiers      & 1$\times$1    & $N$   & \begin{tabular}[c]{@{}l@{}}$N$ is the output size of the classifier. \end{tabular} \\ \hline
\end{tabularx}
}
\end{table}

\subsubsection{Architecture of SCR-Net} 
Taking efficiency into account, we perform a forward-pass CNN rather than use character segmentation or RNNs. The architecture of SCR-Net is shown in Fig. \ref{Fig3}. The detailed descriptions of SCR-Net are shown in Table \ref{table2}. We employ the \textit{BN-Conv3-BN-ReLU-Conv3-BN-Dynamic} \cite{wang2020convolutional} structure in the residual branch of S-Stage-1, 2, and 3, which effectively decreases the overfitting. Moreover, to specialize in the left-to-right format of LPs, we propose a horizontal encoding technique to squeeze features maps. In particular, we design the S-Conv-2 and S-Stage-4 structures. In S-Conv-2, the output feature maps of S-Stage-3 (with the size of 8$\times$32) are encoded by an 8x1-s1 convolutional filter, compressing the feature maps to the size of 1$\times$32. In S-Stage-4, we adopt two residual blocks \cite{he2016deep} followed by the Conv-BN-ReLU to extract LP features where all convolutional filters use a 1$\times$3 kernel. By doing so, the feature maps are vertically squeezed and encoded to 1D horizontal features with the size of 1$\times$32, fitting the left-to-right format.

\begin{figure} [t]
\centering 
\includegraphics[width=8.5cm]{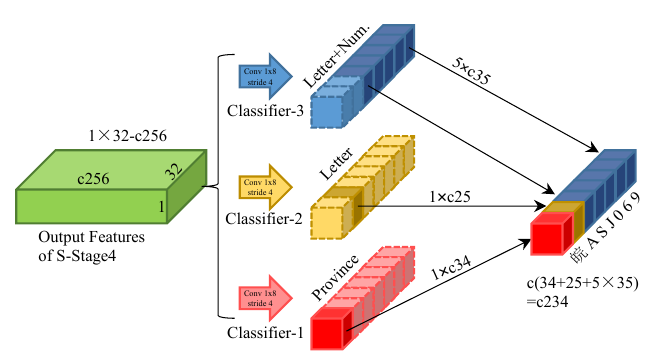}\\ 
\caption{Implementation of the weight-sharing classifiers.}
\label{Fig5} 
\end{figure}

Moreover, a weight-sharing classifier is proposed to balance training samples and address the issue of small-scale datasets, such as AOLP \cite{hsu2012application}. Usually, CNN-based models employ fully-connected layers as a classifier to recognize the character on each predefined position of LPs \cite{xu2018towards,vspavnhel2017holistic}. However, this manner cannot handle small-scale datasets since the probability of a specific character appearing in a specific position is low, and even the character never appears in that position. Our weight-sharing classifier utilizes the shared weights of convolutional layers such that the classifier can spot all instances of characters in the dataset. Fig. \ref{Fig5} shows three weight-sharing classifiers for three types of characters on the CCPD \cite{xu2018towards} dataset. The horizontal features are fed to the weight-sharing classifier, which produces probabilities of each character. Each classifier is implemented by a convolutional layer with the kernel size of 1$\times$8 and the stride of 1$\times$4. For small-scale datasets, like AOLP \cite{hsu2012application}, only one classifier (kernel of 1$\times$7 and stride of 1$\times$5) can handle six-position characters. This structure also decreases parameter consumption compared with fully-connected layers (see Section \ref{abl-scrnet}). Meanwhile, the weight-sharing classifier avoids applying additional character segmentation or RNNs for the small-scale datasets.

\textbf{SCR-Net-Fast.} Like VertexNet-Fast, we also propose a fast version, named SCR-Net-Fast. Specifically, referring to Table \ref{table2}, we change the number of channels of S-Stage-1, 2, 3, S-Conv-2, and S-Stage-4 to 53, 91, 128, 128, and 128, respectively. Besides, we change 4 blocks in S-Stage-1, 2, and 3 to 2 blocks. This shallow architecture improves inference speed with a little performance decrease (see \ref{comp-recognition}).

\subsubsection{Training of SCR-Net} 
We generate pairs of training examples according to the labelled LP vertices and the corresponding characters. The LP image is normalized to the size of 64$\times$256. We predefine the output layout for CCPD \cite{xu2018towards} and AOLP \cite{hsu2012application} datasets. For CCPD, there are three character types arranged on seven places in order, i.e., 34 Chinese characters (provinces) and 25 English characters (letters) in the first and second places of LPs, respectively, followed by 35 English+Number characters in the last five places. Therefore, the output size of the classifier ($N$ in Table \ref{table2}) is $34+25+5\times 35=234$, as shown in Fig. \ref{Fig5}. Similarly, we define 35 English+Number characters arranged on six places for AOLP, so the output size is $210$.

The cross-entropy loss is employed for predicting the characters for each character type, i.e., 
\begin{equation}
\label{eqn6}
L = - y_{i} \cdot \log(F(x_{i},\Omega)),
\end{equation}where $y_{i}$ is the ground-truth one-hot vector, $F(x_{i},\Omega)$ is the prediction of SCR-Net, and ``$\cdot$'' is the inner product. In $F(x_{i},\Omega)$, $x_{i}$ is an input image and $\Omega$ is a set of parameters.

\section{Experiments}\label{Experiment} 

In this section, we first introduce datasets, evaluation metrics, and implementation details. Then, we evaluate the proposed VSNet compared with state-of-the-art methods in terms of LP detection and LP recognition. We conduct ablation studies to investigate the effectiveness of the proposed modules. 

\subsection{Datasets and Evaluation Metrics}

\subsubsection{Dataset}
We evaluate the ALPR methods on four real-world datasets, i.e., CCPD \cite{xu2018towards}, AOLP \cite{hsu2012application}, PKUData \cite{yuan2016robust}, and CLPD \cite{zhang2020robust}.

\textbf{CCPD.} Chinese city parking dataset (CCPD) \cite{xu2018towards} is an extensive and comprehensive LP benchmark to evaluate ALPR methods in uncontrolled conditions. CCPD contains over 280k vehicle images captured in uncontrolled conditions, e.g., different weathers, illuminations, rotation, and vagueness, which is two orders of magnitude more than other LP datasets. Each image has the resolution of $720\times 1160$. The dataset provides sufficient annotations, such as LP character, bounding box, four vertices, horizontal and vertical tilt degree, and the degree of brightness and vagueness. Following \cite{xu2018towards,zhang2020robust}, we train our model on 100k examples of CCPD-Base and test it on the rest 100k examples of CCPD-Base and the 80k examples of sub-datasets including CCPD-DB, CCPD-FN, CCPD-Rotate, CCPD-Tilt, CCPD-Weather, and CCPD-Challenge. 

\textbf{AOLP.} Application-oriented license plate dataset (AOLP) proposed by Hsu \textit{et al.} \cite{hsu2012application} shows three types of constrained subsets, i.e., access control (AC) for the entrance/exit of a region, law enforcement (LE) for the vehicles that violate traffic laws captured by a roadside camera, and road patrol (RP) for the vehicles with arbitrary viewpoints and distances. The AOLP contains 2,049 images, which is challenging for fully connected-based classifiers with many parameters. Like \cite{li2018toward,zhang2020robust}, we train our model on any two subsets and test on the remaining one. We compared our method with state-of-the-art methods in the LP recognition task.

\textbf{PKUData.} Yuan \textit{et al.} \cite{yuan2016robust} introduced an LP detection dataset, where the characters in 2253 images are labeled by \cite{zhang2020robust}. Those images are selected from three subsets: G1 (daytime under normal conditions), G2 (daytime with sunshine glare), and G3 (nighttime). Following the experiments of \cite{zou2020robust}, we evaluate the proposed method's generalization ability for this dataset without training on it.

\textbf{CLPD.} This dataset was introduced by Zhang \textit{et al.} \cite{zhang2020robust}, where 1200 images were collected from all provinces in mainland China. CLPD is also a real-world dataset with a wide variety of environments, vehicle types, and region codes, but it has a different region code distribution compared with CCPD. We follow the setting of \cite{zou2020robust} to test all images. Like PKUData, the test on CLPD shows the practicality of the LP recognition methods. SCR-Net uses a single weight-sharing classifier for PKUData and CLPD.

\subsubsection{Evaluation Metrics}
\label{Metrics}
For the CCPD dataset, we follow the detection and recognition metrics in \cite{xu2018towards}, where the detection precision is calculated without considering the recall since each image only produces one single license plate. The detector is allowed to predict one bounding box for each image. The predicted box is true positive when its IoU with ground truth is larger than 0.7. We report the precision over the testing set (overall precision) and each subset. In the LP recognition task, the predicted LP is correct when its IoU is more than 0.6, and all characters on the LP are correct. The recognition accuracy is calculated by the correct LP predictions over all LPs. Similarly, for the AOLP, PKUData, and CLPD datasets, we calculate the recognition accuracy based on the percentage of true positives. The true positive is defined by those where all characters are correct in an LP. We follow \cite{zhang2020robust} to test the inference time per image on an NVIDIA GTX1080Ti GPU.

\subsection{Implementation Details}
Since VertexNet and SCR-Net employ the proposed architecture, there are no off-the-shelf pre-trained weights on ImageNet, so we trained them from scratch. We trained the VertexNet by 30 epochs with the batch size of 128 in one GPU, which costs only 4 hours. The learning rate was initialized to 0 and warmed up to 0.08 at the first 2000 steps. Then, we applied the cosine learning schedule to gradually reduce the learning rate to 0 at the end of the training. We used the SGD optimizer with the weight decay of 0.0005 and the momentum of 0.9. All input images of VertexNet were rescaled to 256 pixels with the aspect ratio of original images. We employed vertex-based data augmentation, randomly flipping, and AutoAugment \cite{cubuk2019autoaugment}. The vertex-based data augmentation is an affine transformation, expressed as the rotation, translation, and scale operation of input images and LPs' vertices. The affine transformation is implemented by OpenCV.

For training SCR-Net, we employed the SGD optimizer with the momentum of 0.9. The learning rate was initialized to 0 and warmed up to 0.6 at the first 2000 steps. The cosine learning schedule was used to decrease the learning rate during training. We set the training epoch to 100 and the batch size to 128. The dynamic regularization \cite{wang2020convolutional} was applied to each block of S-Stage-1, 2, and 3 (see Table \ref{table2}). In order to against angle variation, we augment an LP patch with the 3D transformation by vertices. We randomly enlarge the sampling width to \(\overline{w}_{l} \in [w_{l}, 1.5\times w_{l}]\) and the sampling height to \(\overline{h}_{l} \in [h_{l}, 1.5 \times h_{l}]\), where \(w_{l}\) and \(h_{l}\) denote the width and height of an LP, respectively. This operation makes SCR-Net robust to the imprecise LP detections of VertexNet. Then, the sampled LP images were resized to 64-pixel height and 256-pixel width as the inputs of SCR-Net. We used AutoAugment \cite{cubuk2019autoaugment} to diverse LP images. The results of VertexNet and SCR-Net were obtained by the mean of three runs.

\begin{table*}[ht]
\centering \caption{Comparisons of LP detection methods on the CCPD testing set. ``-'' means the result is not provided.}
\label{table3}
\begin{tabular}{lccccccccc}
\hline
Method                                                                   & \begin{tabular}[c]{@{}c@{}}Overall\\ Precision\end{tabular}             & \begin{tabular}[c]{@{}c@{}}Base\\ (100k)\end{tabular} & \begin{tabular}[c]{@{}c@{}}DB\\ (20k)\end{tabular} & \begin{tabular}[c]{@{}c@{}}FN\\ (20k)\end{tabular} & \begin{tabular}[c]{@{}c@{}}Rotate\\ (10k)\end{tabular} & \begin{tabular}[c]{@{}c@{}}Title\\ (10k)\end{tabular} & \begin{tabular}[c]{@{}c@{}}Weather\\ (10k)\end{tabular} & \begin{tabular}[c]{@{}c@{}}Challenge\\ (10k)\end{tabular} & \begin{tabular}[c]{@{}c@{}}Inference Time\\ ms/image\end{tabular} \\ \hline
Faster RCNN \cite{ren2015faster}                                                              & 92.9                 & 98.1                                                  & 92.1                                               & 83.7                                               & 91.8                                                   & 89.4                                                  & 81.8                                                    & 83.9                                                      & 56.7                                                     \\
YOLO9000 \cite{redmon2017yolo9000}                                                                & 93.1                 & 98.8                                                  & 89.6                                               & 77.3                                               & 93.3                                                   & 91.8                                                  & 84.2                                                    & 88.6                                                      & 22.8                                                     \\
SSD300 \cite{liu2016ssd}                                                                   & 94.4                 & 99.1                                                  & 89.2                                               & 84.7                                               & 95.6                                                   & 94.9                                                  & 83.4                                                    & 93.1                                                      & 24.6                                                     \\
TE2E \cite{li2018toward}                                                                  & 94.2                 & 98.5                                                  & 91.7                                               & 83.8                                               & 95.1                                                   & 94.5                                                  & 83.6                                                    & 93.1                                                      & 310                                                    \\
RPnet \cite{xu2018towards}                                                                 & 94.5                 & 99.3                                                  & 89.5                                               & 85.3                                               & 94.7                                                   & 93.2                                                  & 84.1                                                    & 92.8                                                      & 11.7                                                     \\
MTLPR \cite{wang2019light}                                                   & 97.7                 & -                                                     & -                                                  & -                                                  & -                                                      & -                                                     & -                                                       & -                                                         & 15.4                                                     \\ \hline
VertexNet-Fast (ours)                                                          & 98.8                & 99.2                                                  & 99.1                                               & \textbf{97.0}                                               & \textbf{99.9}                                                   & 99.6                                                  & 99.7                                                    & 95.1                                                      & \textbf{3.4}                                                     \\
VertexNet (ours)                                                          & \textbf{99.1}        & \textbf{99.6}                                         & \textbf{99.4}                                      & \textbf{97.0}                                      & \textbf{99.9}                                          & \textbf{99.8}                                         & \textbf{99.9}                                           & \textbf{95.6}                                             &  5.7                                                      \\ \hline
\end{tabular}
\end{table*}

\subsection{License Plate Detection}

\subsubsection{Comparison with state-of-the-art LP detection methods}
We first compare the proposed VertexNet with state-of-the-art LP detection methods. Since the CCPD \cite{xu2018towards} dataset contains diverse environments, the results on CCPD can be considered as a key indicator of LP detection in unconstrained environments. The precision is calculated based on a high IoU threshold of 0.7, which requires the detector to produce high-quality bounding boxes. 

The results are reported in Table \ref{table3}, which shows that many generic object detectors, such as SSD300 \cite{liu2016ssd}, YOLO9000 \cite{redmon2017yolo9000}, and Faster RCNN \cite{ren2015faster}, generally perform worse than the LP-specific detection methods, such as TE2E\cite{li2018toward}, PRnet \cite{xu2018towards}, MTLPR \cite{wang2019light}, and VertexNet. Among LP-specific methods, the unified frameworks, TE2E and PRnet, cannot achieve top performance. Due to the detection-oriented optimization and well-designed architecture, the detectors of the cascaded frameworks, MTLPR and VertexNet, outperforms TE2E and PRnet. The proposed VertexNet improves the precision by 1.4\% compared with the state-of-the-art MTLPR, achieving the best precision (99.1\%). Our method obtains the best results on all subsets, especially for CCPD-FN and CCPD-Weather with more than 10\% improvement compared with RPnet. Note that 1\% improvement means that the method can correct 1,800 false detections over 180k testing images. VertexNet runs at 5.7 ms per image. Moreover, with naive IB, VertexNet-Fast obtains the fastest speed at 3.4 ms per image, and its 98.8\% accuracy is also better than state-of-the-art methods.

\begin{figure} [t]
\centering 
\includegraphics[width=8.5cm]{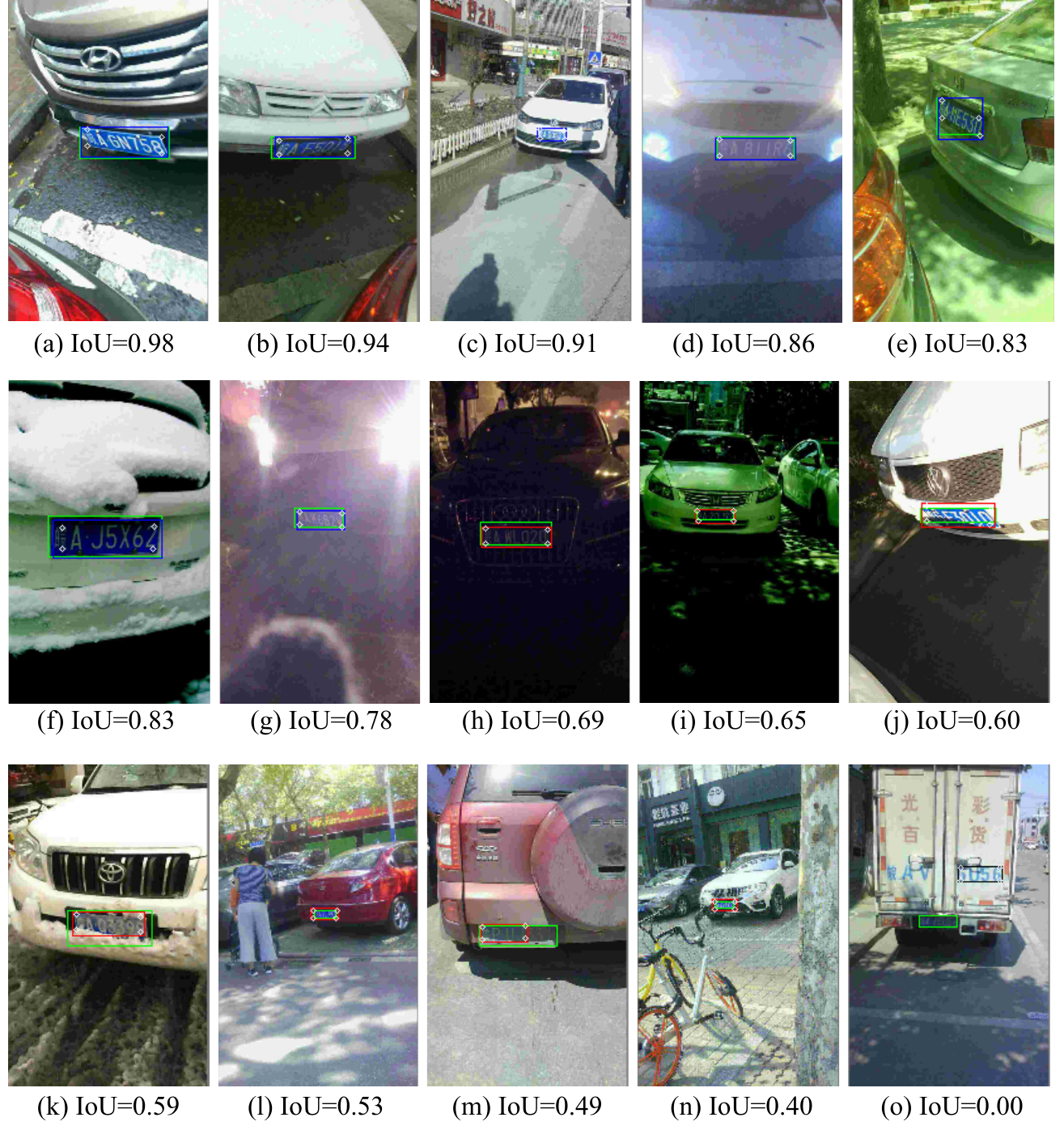}\\ 
\caption{Qualitative results of VertexNet on the CCPD testing set. Green, blue, and red bounding boxes represent ground truth, truth positive detections, and failure detections, respectively. White points denote the predicted vertices. (a)-(g) show truth positive detections with the IoU $\in [0.7, 1.0]$. (h)-(o) shows failure detections with the IoU $\in [0.0, 0.7)$.}
\label{Fig6}
\end{figure}

\begin{figure} [t]
\centering 
\includegraphics[width=9cm]{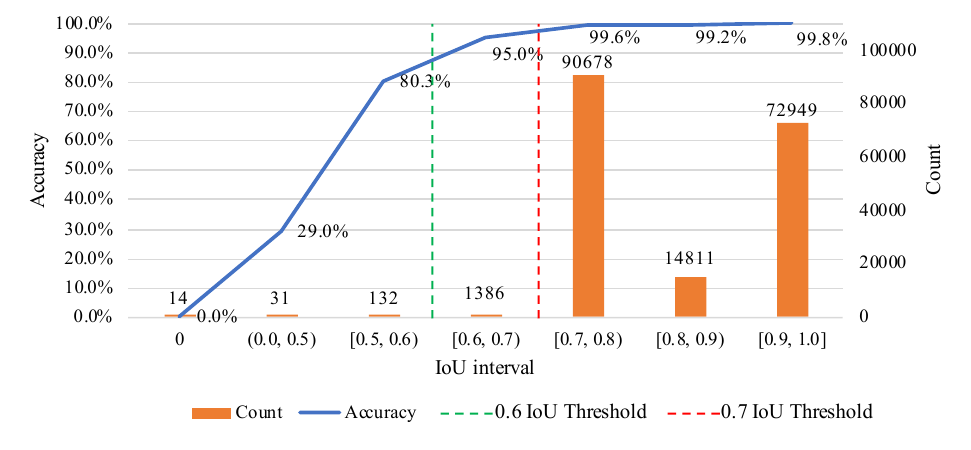}\\ 
\caption{The accuracy of SCR-Net and the count of detections of VertexNet with respect to IoU interval.}
\label{Fig_iou}
\end{figure}

We show some qualitative results of VertexNet on the CCPD testing set in Fig. \ref{Fig6}. We point out that VertexNet is robust for unconstrained environments, such as uneven light and oblique views (see the (a)-(g) of Fig. \ref{Fig6}). VertexNet only fails in 1,563 images over 180k testing images, i.e., 0.9\% error rate, suppressing 2.3\% error rate of MTLPR by a relative improvement of 60\%. Some failure cases are illustrated in the (h)-(o) of Fig. \ref{Fig6}. The failure cases are divided into three aspects: 1) detections with IoU $\in [0.6, 0.7)$, 2) detections with IoU $\in [0.5,0.6)$, and 3) detections with IoU $\in [0.0,0.5)$, and the number of them is 1,386, 132, and 45, respectively, as shown in Fig. \ref{Fig_iou}. Most of the failure cases have the IoU between 0.6 and 0.7. If we set the IoU threshold to 0.6 (the green line of Fig. \ref{Fig_iou}) when calculating precision, the precision of VertexNet can achieve 99.9\%. In LP recognition, we will show that most characters of the failure cases can be still recognized by SCR-Net (see Section \ref{comp-recognition}). 

\begin{table}[t]
\centering \caption{Ablation study of VertexNet on the CCPD testing set. ``Augment'': vertex-based data augmentation. ``Est.'': vertex estimation. ``\#Params'': the number of network's parameters. ``$\surd$'' means using the corresponding technique.}
\label{table4}
\resizebox{\columnwidth}{!}{%
\begin{tabular}{lclccc}
\hline
Backbone                   & Augment & IB setting   & Est.    & Precision   & \#Params \\ \hline
ResNet-50                  &  &      -        &  & 97.5 & 27.2M    \\ \hline
\multirow{5}{*}{IB (ours)} &         & w/o p-norm SE &         & 97.4 & 3.19M     \\
                           &     & w/ SE \cite{hu2018squeeze} &          & 97.5 & 3.27M     \\
                           &     & w/ p-norm SE         &         & 98.3 & 3.31M     \\
                           & $\surd$ & w/ p-norm SE  &         & 98.8 & 3.31M     \\
                           & $\surd$ & w/ p-norm SE  & $\surd$ & 99.1 & 3.48M     \\ \hline
\end{tabular}
}
\end{table}

\begin{table*}[ht]
\centering \caption{Comparisons of LP recognition methods on the CCPD testing set. ``-'' means the result is not provided. ``HC'' is a CNN followed by a fully-connected classifier running at 1 ms per image. SCR-Net is fed by the LP images sampled by the vertices of VertexNet. ``Inference time'' means the whole inference time of VertexNet and SCR-Net. The time in the parentheses means the inference time of SCR-Net only.}
\label{table5}
\begin{tabular}{lccccccccc}
\hline
Method          & \begin{tabular}[c]{@{}c@{}}Overall\\ Accuracy\end{tabular} & \begin{tabular}[c]{@{}c@{}}Base\\ (100k)\end{tabular} & \begin{tabular}[c]{@{}c@{}}DB\\ (20k)\end{tabular} & \begin{tabular}[c]{@{}c@{}}FN\\ (20k)\end{tabular} & \begin{tabular}[c]{@{}c@{}}Rotate\\ (10k)\end{tabular} & \begin{tabular}[c]{@{}c@{}}Title\\ (10k)\end{tabular} & \begin{tabular}[c]{@{}c@{}}Weather\\ (10k)\end{tabular} & \begin{tabular}[c]{@{}c@{}}Challenge\\ (10k)\end{tabular} & \begin{tabular}[c]{@{}c@{}}Inference Time\\ ms/image\end{tabular} \\ \hline
Faster RCNN \cite{ren2015faster} + HC \cite{vspavnhel2017holistic} & 92.8    & 97.2                                                  & 94.4                                               & 90.9                                               & 82.9                                                   & 87.3                                                  & 85.5                                                    & 76.3                                                      & 57.6                                                       \\
YOLO9000 \cite{redmon2017yolo9000} + HC \cite{vspavnhel2017holistic}  & 93.7    & 98.1                                                  & 96.0                                               & 88.2                                               & 84.5                                                   & 88.5                                                  & 87.0                                                    & 80.5                                                      & 23.8                                                       \\
SSD300 \cite{liu2016ssd} + HC \cite{vspavnhel2017holistic}     & 95.2    & 98.3                                                  & 96.6                                               & 95.9                                               & 88.4                                                   & 91.5                                                  & 87.3                                                    & 83.8                                                      & 25.6                                                       \\
TE2E \cite{li2018toward}            & 94.4    & 97.8                                                  & 94.8                                               & 94.5                                               & 87.9                                                   & 92.1                                                  & 86.8                                                    & 81.2                                                      & 310                                                       \\
RPnet \cite{xu2018towards}          & 95.5    & 98.5                                                  & 96.9                                               & 94.3                                               & 90.8                                                   & 92.5                                                  & 87.9                                                    & 85.1                                                      & 11.7                                                       \\
DAN \cite{DAN_aaai20}           & 96.6    & 98.9                                                     & 96.1                                                  & 96.4                                                  & 91.9                                                      & 93.7                                                     & 95.4                                                       & 83.1                                                         & 19.3                                                       \\
Zou \textit{et al.} \cite{zou2020robust}           & 97.8    & 99.3                       & 98.5                                                  & 98.6                                                 & 92.5                                                     & 96.4                                                & 99.3                                                  & 86.6                                                 & -                                               \\
MORAN \cite{luo2019moran}           & 98.3            & 99.5                                & 98.1                                                  & 98.6                                                  & 98.1                                                      & 98.6                                                     & 97.6                                                       & 86.5                                                         & 18.2                                                       \\
Attentional Net \cite{zhang2020robust} & 98.5    & 99.6                                                  & 98.8                                               & 98.8                                               & 96.4                                                   & 97.6                                                  & 98.5                                                    & 88.9                                                      & 24.9                                                       \\
MTLPR \cite{wang2019light}          & 98.8    & -                                                     & -                                                  & -                                                  & -                                                      & -                                                     & -                                                       & -                                                         & 15.6                                                       \\
Attentional Net w/ synthetic data \cite{zhang2020robust} & 98.9    & 99.8                                                  & 99.2                                               & 99.1                                               & 98.1                                                   & 98.8                                                  & 98.6                                                    & 89.7                                                      & 24.9                                                       \\ \hline
SCR-Net-Fast (ours) & 99.4    & \textbf{99.9}                                                  & 99.6                                               & 99.3                                               & 99.8                                                   & 99.8                                                 & 99.1                                                    & 93.9                                                     & \textbf{6.7} (3.3)                                                       \\
SCR-Net (ours)  & \textbf{99.5}    & \textbf{99.9}                                                      & \textbf{99.7}                                                   & \textbf{99.4}                                                   & \textbf{99.9}                                                       & \textbf{99.9}                                                      & \textbf{99.4}                                                        & \textbf{94.8}                                                         & 11.4 (5.7)                                                       \\ \hline
\end{tabular}
\end{table*}

\subsubsection{Ablation study of VertexNet}
\label{abl-vertexnet}
We conduct ablation studies on the CCPD dataset to evaluate the effectiveness of the proposed modules in VertexNet. The modules include IB, vertex-based augmentation, and vertex estimation. Table \ref{table4} shows the results of VertexNet with different settings.

\textbf{Integration block.} As the building block of the backbone network, we first evaluate the effectiveness of the IB by three settings, as shown in the 2-4th entries of Table \ref{table4}. We also set up a strong baseline backbone, ResNet-50 \cite{he2016deep}, as shown in the 1st entry. ResNet-50 obtains 97.5\% precision, but the multiple channels in top layers lead to a large number of parameters. From the 2nd entry of Table \ref{table4}, the IB without the $p$-norm SE that employs the narrow channels and shallow structure is comparable to ResNet-50 (97.4 \textit{vs.} 97.5), which shows that the multiple channels are unnecessary for small object detection. With the SE \cite{hu2018squeeze} module (the 3rd entry), the performance of the IB improves slightly. Furthermore, the IB with the $p$-norm SE (the 4th entry), i.e., the whole structure of the IB, achieves 98.3\% precision. This version outperforms ResNet-50 by a significant margin of 0.8\% and overtakes the state-of-the-art MTLPR \cite{wang2019light}.

\textbf{Vertex-based data augmentation.} Vertex-based data augmentation is to generate diverse views of LPs by performing the affine transformation on input images and LP's vertices. From the 4th and 5th entries of Table \ref{table4}, we can observe that this component boosts the detection precision from 98.3\% to 98.8\%. The improvement illustrates that the vertex-based augmentation is helpful for LP detection.

\textbf{Vertex estimation.} We compare VertexNet without/with the vertex estimation branch, as shown in the last two entries of Table \ref{table4}. With only 5\% relative parameter increases (0.17M/3.31M), the vertex estimation branch improves the detection precision from 98.8\% to 99.1\%. Moreover, the estimated vertices are crucial for the LP rectification in SCR-Net. The rectified LPs can significantly enhance the recognition accuracy. See Section \ref{abl-scrnet} for more detailed comparisons.

\begin{figure} [t]
\centering 
\includegraphics[width=8.5cm]{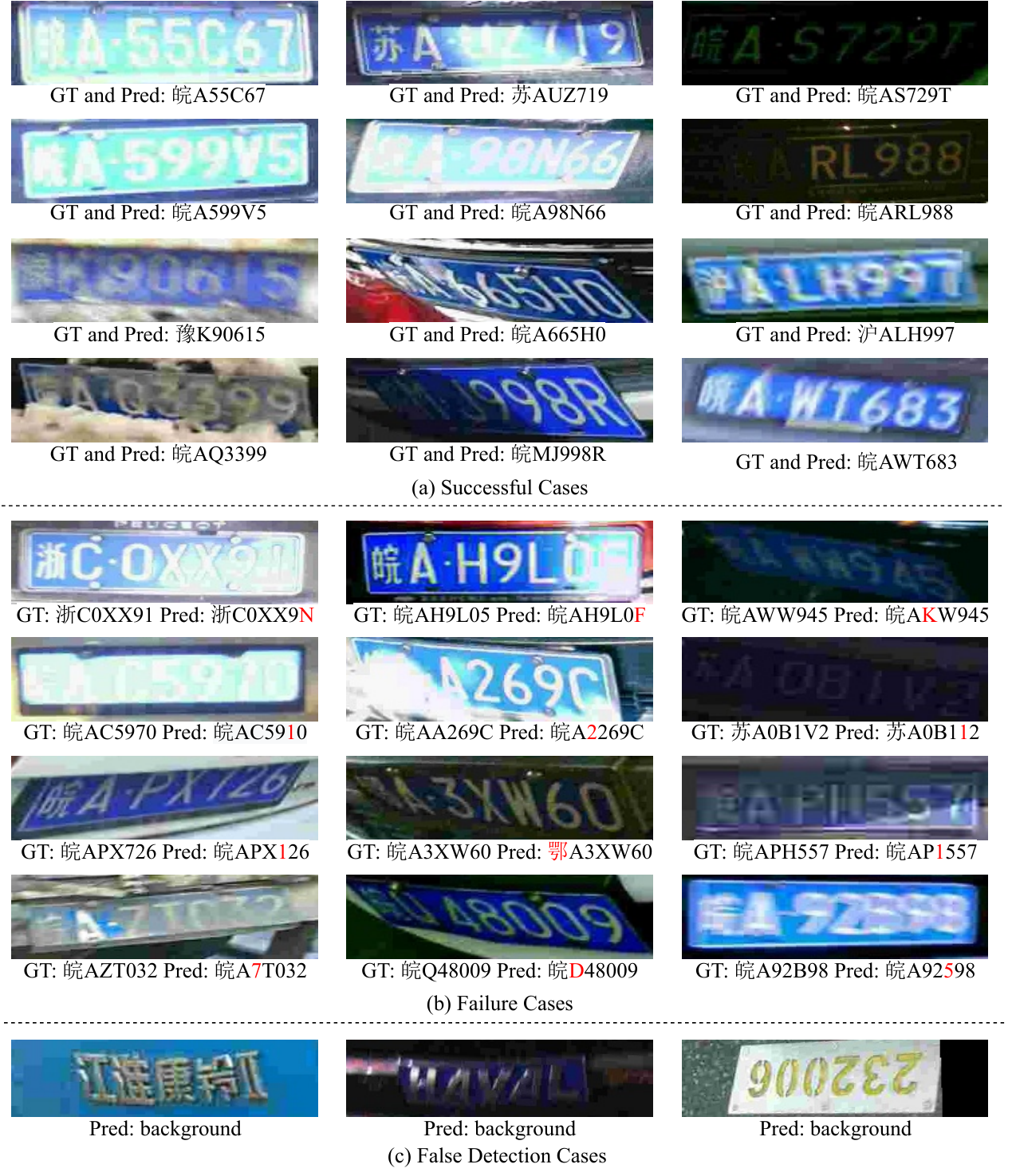}\\ 
\caption{Qualitative results of our SCR-Net. GT: ground truth, and Pred: network prediction. Error characters are marked by the red color. False detections in (c) cause false recognition compared with GT, but SCR-Net can still recognize these images as ``background'' rather than LPs.}
\label{Fig7}
\end{figure}

\subsection{License Plate Recognition}

\subsubsection{Comparison with state-of-the-art LP recognition methods}
\label{comp-recognition}
We first compare the proposed SCR-Net with state-of-the-art LP recognition methods on the CCPD \cite{xu2018towards} and AOLP \cite{hsu2012application} datasets. Then, we test the CCPD-trained SCR-Net on the other real-world datasets (PKUData \cite{yuan2016robust} and CLPD \cite{zhang2020robust}) to validate the generalization ability of our method.

\textbf{Results on CCPD.} Table \ref{table5} shows the comparisons of LP recognition methods on the CCPD dataset. Instead of using ground-truth vertices to sample LP images, the proposed SCR-Net takes the predicted vertices by VertexNet as the inputs, i.e., VSNet (the cascaded framework of VertexNet and SCR-Net). The results show that SCR-Net is superior to state-of-the-art LP recognition methods, including RNN-based MTLPR \cite{wang2019light} and Attentional Net \cite{zhang2020robust}. SCR-Net obtains the best recognition accuracy (99.5\%) on the overall testing set and all subsets, especially for CCPD-Challenge with more than 5\% accuracy improvement. Our approach are even better than Attentional Net trained with additional synthetic data (99.5\% \textit{vs.} 98.9\%). For inference time, the fast version of our approach (SCR-Net-Fast) obtains the fastest inference speed at 6.7 ms per image, 43\% relative improvement compared with 11.7 ms per image of RPnet \cite{xu2018towards}. This improvement comes from the resampling-based cascaded framework and compact network architectures. 

We detail the result of SCR-Net by partitioning the IoU of detections. The accuracy of SCR-Net and the count of detections of VertexNet with respect to IoU interval are shown in Fig. \ref{Fig_iou}. It can be observed that 1) more than 99\% LPs can be recognized when the IoU is large than 0.7, 2) SCR-Net can obtain 95\% recognition accuracy for IoU $\in [0.6, 0.7)$, and 3) SCR-Net can still correctly recognize some LPs in the case of IoU $< 0.6$. 

\begin{table}[t]
\centering \caption{Comparisons of LP recognition methods on the AOLP dataset. ``GT LPs'' means taking the ground-truth LPs as input. ``box'' means taking the box detection predicted by Attentional Net \cite{zhang2020robust} as input. ``vertex'' is taking the ground truth vertex as input. ``FC'' is the fully-connected classifier.}
\label{table6}
\begin{tabular}{lcccc}
\hline
Method          & Overall & AC    & LE    & RP    \\
\#Images         & (2049)  & (681) & (757) & (611) \\ \hline
TE2E \cite{li2018toward}            & 94.5       & 95.3  & 96.6  & 83.7  \\
Attentional Net \cite{zhang2020robust} & 96.1        & 97.3  & 98.3  & 91.9 \\
Zou \textit{et al.} \cite{zou2020robust} & 95.8 & 97.1  & 96.6  & 93.4 \\ 
Zou \textit{et al.} \cite{zou2020robust} (GT LPs) & 97.8 & 99.3  & 98.7  & 95.1 \\\hline
SCR-Net (box)  & 98.7        & 97.9      &  98.9     &  99.5     \\ 
SCR-Net (vertex)  & \textbf{99.7}        & \textbf{99.4}      &  \textbf{99.9}     &  \textbf{99.7}  \\ \hline
SCR-Net w/ FC  & 81.3        & 97.1     &  53.9    &  97.6 \\ \hline
\end{tabular}
\end{table}

Some successful and failure cases in challenging environments are presented in Fig. \ref{Fig7}. The environments include strong light, uneven light, nighttime, occlusion, oblique view, and blur. SCR-Net handles most of such conditions and fails in few cases. There are only 864 failure cases over 180k testing images, i.e., 0.5\% error rate, defeating 1.1\% error rate of Attentional Net by a relative improvement of 55\%. Among failure cases, human eyes cannot even recognize some characters, like ``5'' recognized as ``F'' in uneven light. Moreover, SCR-Net can filter out false detections of VertexNet by training randomly sampled background images. The false detections in Fig. \ref{Fig7} (c) are recognized as ``background''.

\textbf{Results on AOLP.} Table \ref{table6} shows the results of LP recognition methods on three subsets of AOLP \cite{hsu2012application}. For a fair comparison, we employed the box detections by \cite{zhang2020robust} as the input images for SCR-Net and did not adopt LP rectification. Our SCR-Net (box) achieves the best accuracy (98.7\%) among state-of-the-art LP recognition methods. Especially on the RP subset, SCR-Net (box) improve the accuracy by 7.6\% when compared with Attentional Net \cite{zhang2020robust}. SCR-Net (box) is still better than Zou \textit{et al.} \cite{zou2020robust} (GT LPs) who takes the ground-truth LPs as inputs. With vertex information, SCR-Net (vertex) improves the accuracy to 99.7\%.

\textbf{Generalization results on PKUData and CLPD.} The generalization of LP recognition algorithms is an ability to recognize unseen LP images collected from different environments. Following the experimental setting of \cite{zhang2020robust,zou2020robust}, we conduct the cross-dataset tests, i.e., we train our SCR-Net with the CCPD \cite{xu2018towards} training set and test it on the PKUData \cite{yuan2016robust} and CLPD \cite{zhang2020robust} datasets. Table \ref{table7} shows the comparisons between the proposed SCR-Net and state-of-the-art LP recognition methods. From the table, the recognition accuracy is presented in two forms, i.e., the accuracy without considering Chinese region characters in the first place of LPs (w/o RC) and the overall accuracy. On CLPD, the overall accuracy of SCR-Net improves to 89.8\% from the 80.7\% accuracy of Zou \textit{et al.}'s \cite{zou2020robust}. The improvement on PKUData is limited because some Chinese characters that appear in the last place of LPs are almost non-existent in the CCPD training set, leading to false recognition for those LPs. SCR-Net can achieve 99.3\% accuracy without considering all Chinese characters on PKUData.

\begin{table}[t]
\centering \caption{Comparisons of the generalization of LP recognition methods on the PKUData and CLPD datasets. ``RC'' is the Chinese region character in the first place of LPs. ``Syn.'' stands for the method trained with the additional synthetic data.}
\label{table7}
\resizebox{\columnwidth}{!}{%
\begin{tabular}{lcccc}
\hline
\multirow{2}{*}{Method}                                                      & \multicolumn{2}{c}{PKUData}   & \multicolumn{2}{c}{CLPD}      \\ \cline{2-5} 
                                                                             & w/o RC        & Overall       & w/o RC        & Overall       \\ \hline
Masood \textit{et al.} \cite{masood2017license}  
& 89.3          & -             & 85.2          & -             \\
RPnet \cite{xu2018towards}                                                                       & 78.4          & 77.6          & 78.9          & 66.5          \\
Attentional Net \cite{zhang2020robust}                                                             & 86.5          & 84.8          & 86.1          & 70.8          \\
Attentional Net w/ Syn. \cite{zhang2020robust}  & 90.5          & 88.2          & 87.6          & 76.8          \\
Zou \textit{et al.} \cite{zou2020robust}       & 96.6          & 96.5          & 94.0          & 80.7          \\ \hline
SCR-Net (ours)                                                                      & \textbf{96.9} & \textbf{96.8} & \textbf{95.3} & \textbf{89.8} \\ \hline
\end{tabular}
}
\end{table}

\begin{figure} [t]
\centering 
\includegraphics[width=9cm]{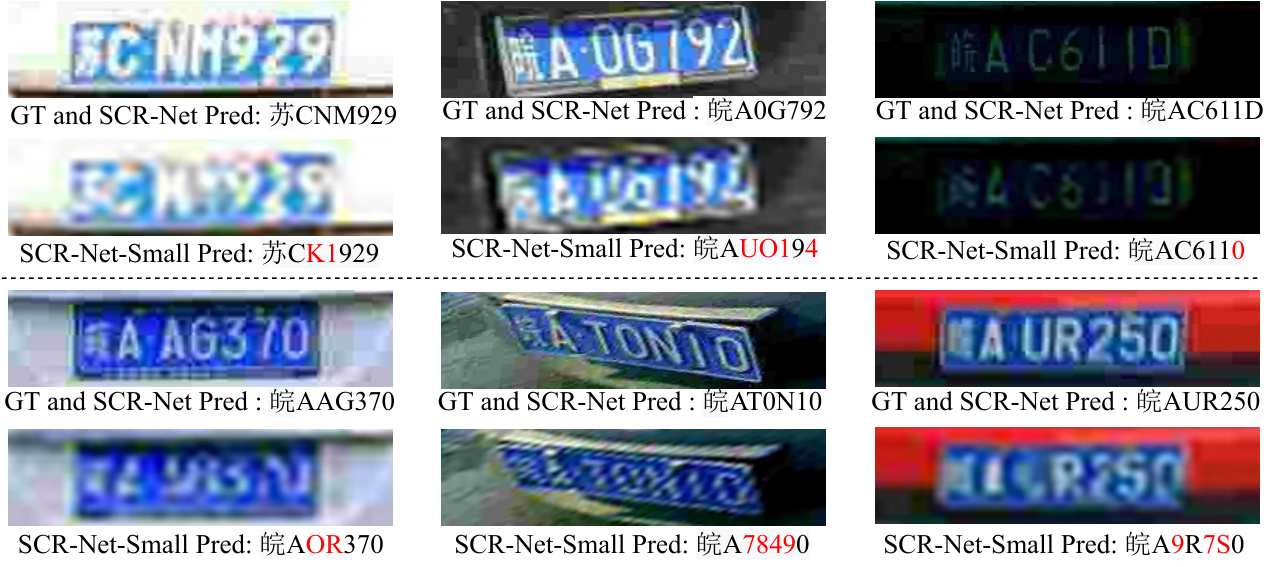}\\ 
\caption{Some LP samples with their predictions by SCR-Net and SCR-Net-Small. GT: ground truth, and Pred: network prediction. Error characters are marked by the red color.}
\label{Fig9}
\end{figure}

\subsubsection{Ablation study of SCR-Net}
\label{abl-scrnet}
We conduct ablation studies to evaluate and analyze the proposed modules in SCR-Net, i.e., resampling-based cascaded framework, vertex-based LP rectification, robustness of SCR-Net, and weight-sharing classifier.

\textbf{Resampling-based cascaded framework.} In TE2E \cite{li2018toward} and RPnet \cite{xu2018towards}, the unified framework only processes one input size. However, the small-size input may lose the details of characters, while the large-size input will increase the inference time. In contrast, our resampling-based cascaded framework decouples the input-size design of VertexNet and SCR-Net, i.e., VertexNet processes small-size images for fast inference regardless of the details of characters, and SCR-Net resamples LP patches from the finest input images. Here we evaluated the importance of the resampling process. If the resampling is disabled (like the unified manner), SCR-Net and VertexNet can only access input images with the size of 256$\times$256, and the LP patches are sampled from those images, named SCR-Net-Small. This setting causes a dramatic drop in the recognition accuracy (94.0\%), compared with 99.5\% accuracy of SCR-Net. Some sampled LP patches with their predictions by SCR-Net and SCR-Net-Small are shown in Fig. \ref{Fig9}. We can observe that due to the small-size input of SCR-Net-Small, the sampled LPs lose the details on some characters. Therefore, SCR-Net-Small makes many mistakes when SCR-Net generates correct predictions. The results illustrates that our resampling process is important to SCR-Net.

\textbf{Vertex-based LP rectification.} LP rectification is crucial for LP recognition in our CNN-based architecture because the rectified LPs imply the character's layout. The more accurate the rectified LPs are, the better the character recognition achieves. The LP rectification is based on the vertex estimation of VertexNet. We compared two types in SCR-Net, which were separately trained by the rectified LPs (resampled LP images by ground-truth vertices) and non-rectified LPs (resampled LP images by ground-truth bounding boxes). In inference, the first model performs LP rectification according to the predicted vertices by VertexNet, while the second one directly sample LPs by the predicted bounding boxes. We evaluated our method on the CCPD testing set. As illustrated in the first two entries (columns) of Table \ref{table8}, the rectified LPs by vertices improve the accuracy from 98.51\% to 99.52\%, which validates the effectiveness of LP rectification. Additionally, the results in Table \ref{table6} show that SCR-Net (vertex) also benefits from the vertices on the AOLP dataset.

\begin{table}[t]
\centering \caption{Ablation Study of SCR-Net with different input types. ``Box'' and ``Vertex'' mean the LPs are sampled according to bounding boxes and vertices, respectively. ``Predicted'' and ``GT'' mean the boxes (or vertices) are generated by VertexNet and ground truth, respectively. ``revised metric'' in the parentheses means removing the condition of IoU $> 0.6$.}
\label{table8}
\resizebox{\columnwidth}{!}{%
\begin{tabular}{ccccc}
\hline
\begin{tabular}[c]{@{}c@{}}Input\\ type\end{tabular} & \begin{tabular}[c]{@{}c@{}} Predicted Box\\(revised metric)\end{tabular} & \begin{tabular}[c]{@{}c@{}}Predicted Vertex\\(revised metric)\end{tabular} & GT Box& GT Vertex\\ \hline
Accuracy                                            & 98.51 (98.56)                                                               & 99.52 (99.58)                                                                  & 98.79     & 99.65        \\ \hline
\end{tabular}
}
\end{table}

\textbf{Robustness of SCR-Net.} SCR-Net may be affected by the biased results of VertexNet. We first modified the recognition metric of CCPD in Section \ref{Metrics} by removing the condition of IoU larger than 0.6. The increase of accuracy in the parentheses of Table \ref{table8} shows that even though the predicted boxes have low IoU ($< 0.6$), SCR-Net can make correct recognition for some inaccurate boxes. Then, we conducted two comparative experiments for the predicted vertices (resp. boxes) and ground-truth vertices (resp. boxes) on the CCPD testing set. From the four entries of Table \ref{table8}, we can observe that the predicted vertices (resp. boxes) only cause $99.65\%-99.58\%=0.07\%$ (resp. $98.79\%-98.56\%=0.23\%$) more errors than the ground-truth vertices (resp. boxes). The results illustrate that SCR-Net is robust to the output of VertexNet, especially for the predicted vertices.

\textbf{Weight-sharing classifier.} The weight-sharing classifier aims to increase training samples for the predefined position of LP images. In previous classification works \cite{xu2018towards,vspavnhel2017holistic}, the fully-connected (FC) layer is employed as a classifier. We compared our classifier with the normal FC layer on the small-scale dataset, i.e., AOLP. Specifically, we replace the weight-sharing classifier with the FC layer. From the last two entries of Table \ref{table6}, we can observe that the weight-sharing classifier is superior to the FC counterpart by a significant margin, i.e., 99.7\% \textit{vs.} 81.1\%. For the LE subset, the sharp drop in performance is caused by the lack of training samples for the FC layer. For example, the ``S'' appears 20 times at the first place of LPs on the LE subset, but there is only one case on the training set, leading to 20 errors in testing. Regardless of the ``S'' position, there are 41 cases in the training set when the weight-sharing classifier is used. With the weight-sharing classifier, our CNN-based model can handle small-scale datasets rather than seek help from RNNs or character segmentation. Moreover, the weight-sharing classifier only requires 4\% parameters of the FC layer (63k \textit{vs.} 1,721k). 

\section{Conclusion}\label{Conclusion} 
This paper has investigated four insights in designing ALPR systems and proposed a high-performing ALPR network, i.e., VSNet. In VSNet, the proposed VertexNet and SCR-Net are connected in a resample-based cascaded manner, improving the inference speed. With a novel architecture design, such as integration block, vertex estimation, horizontal encoding, and weight-sharing classifier, our method achieves the best performance on CCPD and AOLP datasets. Besides, our method performs well in the unseen images of PKUData and CLPD. We validated that taking advantage of vertex information and the weight-sharing classifier is helpful for the CNN model rather than adopt additional character segmentation modules or RNNs. Since the time cost of vertex annotations is equivalent to that of bounding boxes, we think those vertex annotations are more meaningful than box annotations for license plate detection and recognition. We believe our design insights are potential for further research in the ALPR task. For some challenging conditions (like CCPD-Challenge) and unbalanced training data, the self-attention mechanism and generative model can further improve the character recognition performance.


%

\ifCLASSOPTIONcaptionsoff
  \newpage
\fi



%

%
\bibliographystyle{IEEEbib}
\bibliography{ref}

\begin{thebibliography}{10}

\bibitem{anagnostopoulos2006license}
Christos Nikolaos~E Anagnostopoulos, Ioannis~E Anagnostopoulos, Vassilis
  Loumos, and Eleftherios Kayafas,
\newblock ``A license plate-recognition algorithm for intelligent
  transportation system applications,''
\newblock {\em IEEE Transactions on Intelligent Transportation Systems}, vol.
  7, no. 3, pp. 377--392, 2006.

\bibitem{zhao2019object}
Zhong-Qiu Zhao, Peng Zheng, Shou-tao Xu, and Xindong Wu,
\newblock ``Object detection with deep learning: A review,''
\newblock {\em IEEE Transactions on Neural Networks and Learning Systems}, vol.
  30, no. 11, pp. 3212--3232, 2019.

\bibitem{chen2017deeplab}
Liang-Chieh Chen, George Papandreou, Iasonas Kokkinos, Kevin Murphy, and Alan~L
  Yuille,
\newblock ``Deeplab: Semantic image segmentation with deep convolutional nets,
  atrous convolution, and fully connected crfs,''
\newblock {\em IEEE Transactions on Pattern Analysis and Machine Intelligence},
  vol. 40, no. 4, pp. 834--848, 2017.

\bibitem{DAN_aaai20}
Tianwei Wang, Yuanzhi Zhu, Lianwen Jin, Canjie Luo, Xiaoxue Chen, Yaqiang Wu,
  Qianying Wang, and Mingxiang Cai,
\newblock ``Decoupled attention network for text recognition,''
\newblock in {\em AAAI Conference on Artificial Intelligence}, 2020.

\bibitem{luo2019moran}
Canjie Luo, Lianwen Jin, and Zenghui Sun,
\newblock ``Moran: A multi-object rectified attention network for scene text
  recognition,''
\newblock {\em Pattern Recognition}, vol. 90, pp. 109--118, 2019.

\bibitem{li2018toward}
Hui Li, Peng Wang, and Chunhua Shen,
\newblock ``Toward end-to-end car license plate detection and recognition with
  deep neural networks,''
\newblock {\em IEEE Transactions on Intelligent Transportation Systems}, vol.
  20, no. 3, pp. 1126--1136, 2018.

\bibitem{montazzolli2018license}
Sergio Montazzolli~Silva and Claudio Rosito~Jung,
\newblock ``License plate detection and recognition in unconstrained
  scenarios,''
\newblock in {\em Proceedings of the European Conference on Computer Vision},
  2018, pp. 580--596.

\bibitem{ren2015faster}
Shaoqing Ren, Kaiming He, Ross Girshick, and Jian Sun,
\newblock ``Faster r-cnn: Towards real-time object detection with region
  proposal networks,''
\newblock in {\em Advances in Neural Information Processing Systems}, 2015, pp.
  91--99.

\bibitem{xu2018towards}
Zhenbo Xu, Wei Yang, Ajin Meng, Nanxue Lu, Huan Huang, Changchun Ying, and
  Liusheng Huang,
\newblock ``Towards end-to-end license plate detection and recognition: A large
  dataset and baseline,''
\newblock in {\em Proceedings of the European Conference on Computer Vision},
  2018, pp. 255--271.

\bibitem{laroca2019efficient}
Rayson Laroca, Luiz~A. Zanlorensi, Gabriel~R. Gonçalves, Eduardo Todt,
  William~Robson Schwartz, and David Menotti,
\newblock ``An efficient and layout-independent automatic license plate
  recognition system based on the yolo detector,''
\newblock {\em IET Intelligent Transport Systems}, vol. 15, no. 4, pp.
  483--503, 2021.

\bibitem{redmon2017yolo9000}
Joseph Redmon and Ali Farhadi,
\newblock ``Yolo9000: better, faster, stronger,''
\newblock in {\em Proceedings of the IEEE Conference on Computer Vision and
  Pattern Recognition}, 2017, pp. 7263--7271.

\bibitem{silva2017real}
Sergio~Montazzolli Silva and Claudio~Rosito Jung,
\newblock ``Real-time brazilian license plate detection and recognition using
  deep convolutional neural networks,''
\newblock in {\em Conference on Graphics, Patterns and Images (SIBGRAPI)}.
  IEEE, 2017, pp. 55--62.

\bibitem{zhuang2018towards}
Jiafan Zhuang, Saihui Hou, Zilei Wang, and Zheng-Jun Zha,
\newblock ``Towards human-level license plate recognition,''
\newblock in {\em Proceedings of the European Conference on Computer Vision},
  2018, pp. 306--321.

\bibitem{wang2019light}
Wanwei Wang, Jun Yang, Min Chen, and Peng Wang,
\newblock ``A light cnn for end-to-end car license plates detection and
  recognition,''
\newblock {\em IEEE Access}, vol. 7, pp. 173875--173883, 2019.

\bibitem{zhang2020robust}
Linjiang Zhang, Peng Wang, Hui Li, Zhen Li, Chunhua Shen, and Yanning Zhang,
\newblock ``A robust attentional framework for license plate recognition in the
  wild,''
\newblock {\em IEEE Transactions on Intelligent Transportation Systems}, pp.
  1--10, 2020.

\bibitem{zou2020robust}
Yongjie Zou, Yongjun Zhang, Jun Yan, Xiaoxu Jiang, Tengjie Huang, Haisheng Fan,
  and Zhongwei Cui,
\newblock ``A robust license plate recognition model based on bi-lstm,''
\newblock {\em IEEE Access}, vol. 8, pp. 211630--211641, 2020.

\bibitem{hsu2012application}
Gee-Sern Hsu, Jiun-Chang Chen, and Yu-Zu Chung,
\newblock ``Application-oriented license plate recognition,''
\newblock {\em IEEE Transactions on Vehicular Technology}, vol. 62, no. 2, pp.
  552--561, 2012.

\bibitem{yuan2016robust}
Yule Yuan, Wenbin Zou, Yong Zhao, Xinan Wang, Xuefeng Hu, and Nikos Komodakis,
\newblock ``A robust and efficient approach to license plate detection,''
\newblock {\em IEEE Transactions on Image Processing}, vol. 26, no. 3, pp.
  1102--1114, 2016.

\bibitem{du2012automatic}
Shan Du, Mahmoud Ibrahim, Mohamed Shehata, and Wael Badawy,
\newblock ``Automatic license plate recognition (alpr): A state-of-the-art
  review,''
\newblock {\em IEEE Transactions on Circuits and Systems for Video Technology},
  vol. 23, no. 2, pp. 311--325, 2012.

\bibitem{al2018ensemble}
Meeras~Salman Al-Shemarry, Yan Li, and Shahab Abdulla,
\newblock ``Ensemble of adaboost cascades of 3l-lbps classifiers for license
  plates detection with low quality images,''
\newblock {\em Expert Systems with Applications}, vol. 92, pp. 216--235, 2018.

\bibitem{girshick2015fast}
Ross Girshick,
\newblock ``Fast r-cnn,''
\newblock in {\em Proceedings of the IEEE International Conference on Computer
  Vision}, 2015, pp. 1440--1448.

\bibitem{dong2017cnn}
Meng Dong, Dongliang He, Chong Luo, Dong Liu, and Wenjun Zeng,
\newblock ``A cnn-based approach for automatic license plate recognition in the
  wild.,''
\newblock in {\em The British Machine Vision Conference}, 2017.

\bibitem{szegedy2016rethinking}
Christian Szegedy, Vincent Vanhoucke, Sergey Ioffe, Jon Shlens, and Zbigniew
  Wojna,
\newblock ``Rethinking the inception architecture for computer vision,''
\newblock in {\em Proceedings of the IEEE Conference on Computer Vision and
  Pattern Recognition}, 2016, pp. 2818--2826.

\bibitem{liu2016ssd}
Wei Liu, Dragomir Anguelov, Dumitru Erhan, Christian Szegedy, Scott Reed,
  Cheng-Yang Fu, and Alexander~C Berg,
\newblock ``Ssd: Single shot multibox detector,''
\newblock in {\em Proceedings of the European Conference on Computer Vision}.
  Springer, 2016, pp. 21--37.

\bibitem{redmon2016you}
Joseph Redmon, Santosh Divvala, Ross Girshick, and Ali Farhadi,
\newblock ``You only look once: Unified, real-time object detection,''
\newblock in {\em Proceedings of the IEEE Conference on Computer Vision and
  Pattern Recognition}, 2016, pp. 779--788.

\bibitem{gonccalves2016benchmark}
Gabriel~Resende Gon{\c{c}}alves, Sirlene Pio~Gomes da~Silva, David Menotti, and
  William~Robson Schwartz,
\newblock ``Benchmark for license plate character segmentation,''
\newblock {\em Journal of Electronic Imaging}, vol. 25, no. 5, pp. 053034,
  2016.

\bibitem{xie2018new}
Lele Xie, Tasweer Ahmad, Lianwen Jin, Yuliang Liu, and Sheng Zhang,
\newblock ``A new cnn-based method for multi-directional car license plate
  detection,''
\newblock {\em IEEE Transactions on Intelligent Transportation Systems}, vol.
  19, no. 2, pp. 507--517, 2018.

\bibitem{hsu2017robust}
Gee-Sern Hsu, ArulMurugan Ambikapathi, Sheng-Luen Chung, and Cheng-Po Su,
\newblock ``Robust license plate detection in the wild,''
\newblock in {\em IEEE International Conference on Advanced Video and Signal
  Based Surveillance}. IEEE, 2017, pp. 1--6.

\bibitem{laroca2018robust}
Rayson Laroca, Evair Severo, Luiz~A Zanlorensi, Luiz~S Oliveira,
  Gabriel~Resende Gon{\c{c}}alves, William~Robson Schwartz, and David Menotti,
\newblock ``A robust real-time automatic license plate recognition based on the
  yolo detector,''
\newblock in {\em International Joint Conference on Neural Networks}. IEEE,
  2018, pp. 1--10.

\bibitem{chen2019simultaneous}
Song-Lu Chen, Chun Yang, Jia-Wei Ma, Feng Chen, and Xu-Cheng Yin,
\newblock ``Simultaneous end-to-end vehicle and license plate detection with
  multi-branch attention neural network,''
\newblock {\em IEEE Transactions on Intelligent Transportation Systems}, 2019.

\bibitem{qin2016joint}
Hongwei Qin, Junjie Yan, Xiu Li, and Xiaolin Hu,
\newblock ``Joint training of cascaded cnn for face detection,''
\newblock in {\em Proceedings of the IEEE Conference on Computer Vision and
  Pattern Recognition}, 2016, pp. 3456--3465.

\bibitem{gou2015vehicle}
Chao Gou, Kunfeng Wang, Yanjie Yao, and Zhengxi Li,
\newblock ``Vehicle license plate recognition based on extremal regions and
  restricted boltzmann machines,''
\newblock {\em IEEE Transactions on Intelligent Transportation Systems}, vol.
  17, no. 4, pp. 1096--1107, 2015.

\bibitem{graves2008novel}
Alex Graves, Marcus Liwicki, Santiago Fern{\'a}ndez, Roman Bertolami, Horst
  Bunke, and J{\"u}rgen Schmidhuber,
\newblock ``A novel connectionist system for unconstrained handwriting
  recognition,''
\newblock {\em IEEE Transactions on Pattern Analysis and Machine Intelligence},
  vol. 31, no. 5, pp. 855--868, 2008.

\bibitem{he2016deep}
Kaiming He, Xiangyu Zhang, Shaoqing Ren, and Jian Sun,
\newblock ``Deep residual learning for image recognition,''
\newblock in {\em Proceedings of the IEEE Conference on Computer Vision and
  Pattern Recognition}, 2016, pp. 770--778.

\bibitem{hu2018squeeze}
Jie Hu, Li~Shen, and Gang Sun,
\newblock ``Squeeze-and-excitation networks,''
\newblock in {\em Proceedings of the IEEE Conference on Computer Vision and
  Pattern Recognition}, 2018, pp. 7132--7141.

\bibitem{lin2017feature}
Tsung-Yi Lin, Piotr Doll{\'a}r, Ross Girshick, Kaiming He, Bharath Hariharan,
  and Serge Belongie,
\newblock ``Feature pyramid networks for object detection,''
\newblock in {\em Proceedings of the IEEE Conference on Computer Vision and
  Pattern Recognition}, 2017, pp. 2117--2125.

\bibitem{wang2020convolutional}
Yi~Wang, Zhen-Peng Bian, Junhui Hou, and Lap-Pui Chau,
\newblock ``Convolutional neural networks with dynamic regularization,''
\newblock {\em IEEE Transactions on Neural Networks and Learning Systems},
  2020.

\bibitem{vspavnhel2017holistic}
Jakub {\v{S}}pa{\v{n}}hel, Jakub Sochor, Roman Jur{\'a}nek, Adam Herout,
  Luk{\'a}{\v{s}} Mar{\v{s}}{\'\i}k, and Pavel Zem{\v{c}}{\'\i}k,
\newblock ``Holistic recognition of low quality license plates by cnn using
  track annotated data,''
\newblock in {\em IEEE International Conference on Advanced Video and Signal
  Based Surveillance}. IEEE, 2019, pp. 1--6.

\bibitem{cubuk2019autoaugment}
Ekin~D Cubuk, Barret Zoph, Dandelion Mane, Vijay Vasudevan, and Quoc~V Le,
\newblock ``Autoaugment: Learning augmentation strategies from data,''
\newblock in {\em Proceedings of the IEEE Conference on Computer Vision and
  Pattern Recognition}, 2019, pp. 113--123.

\bibitem{masood2017license}
Syed~Zain Masood, Guang Shu, Afshin Dehghan, and Enrique~G Ortiz,
\newblock ``License plate detection and recognition using deeply learned
  convolutional neural networks,''
\newblock {\em arXiv preprint arXiv:1703.07330}, 2017.

\end{thebibliography}


\begin{IEEEbiography}[{\includegraphics[width=1in,height=1.25in,clip,keepaspectratio]{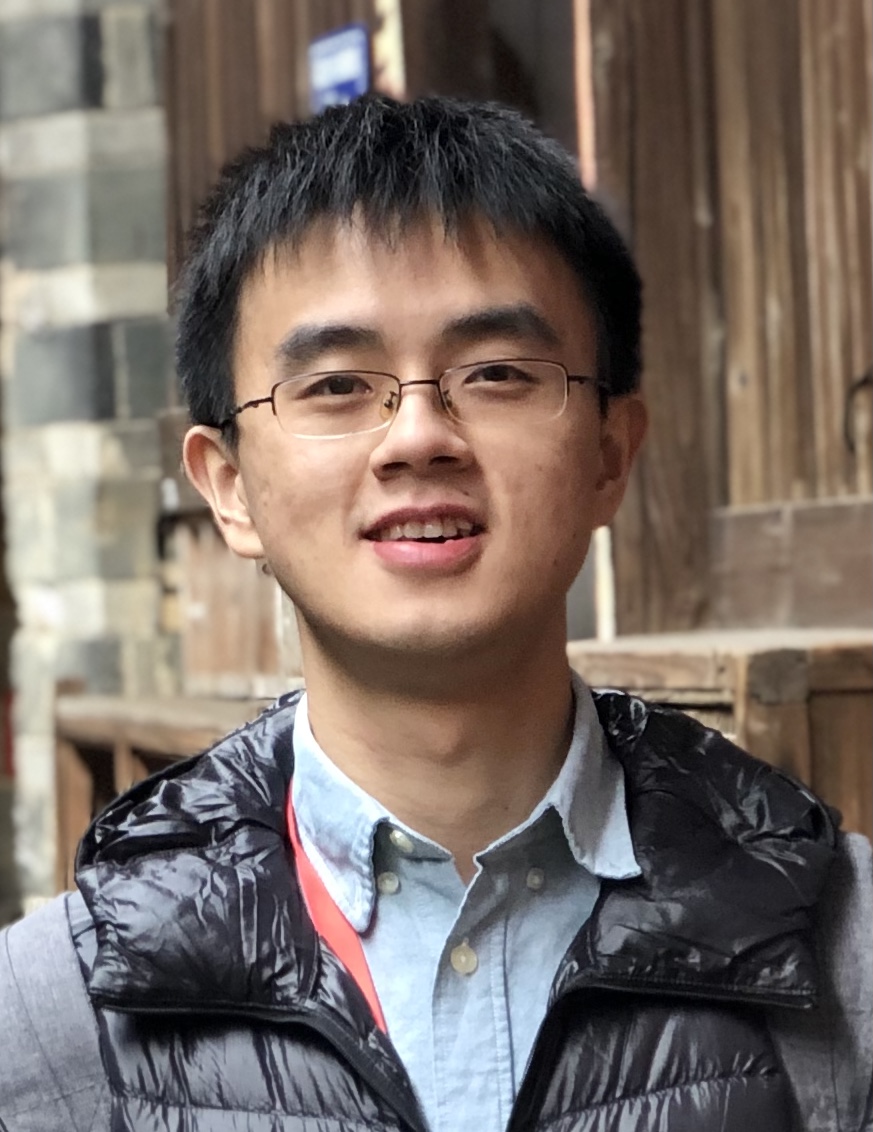}}]{Yi Wang}
received the B.Eng. degree in electronic information engineering and M.Eng. degree in information and signal processing from the School of Electronics and Information, Northwestern Polytechnical University, Xi'an, China, in 2013 and 2016, respectively. He is currently a research associate with the School of Electrical and Electronic Engineering, Nanyang Technological University. He is also working toward the Ph.D. degree of Nanyang Technological University. His research interests include image restoration, image recognition, object detection and tracking, and crowd analysis.
\end{IEEEbiography}

\begin{IEEEbiography}[{\includegraphics[width=1in,height=1.25in,clip,keepaspectratio]{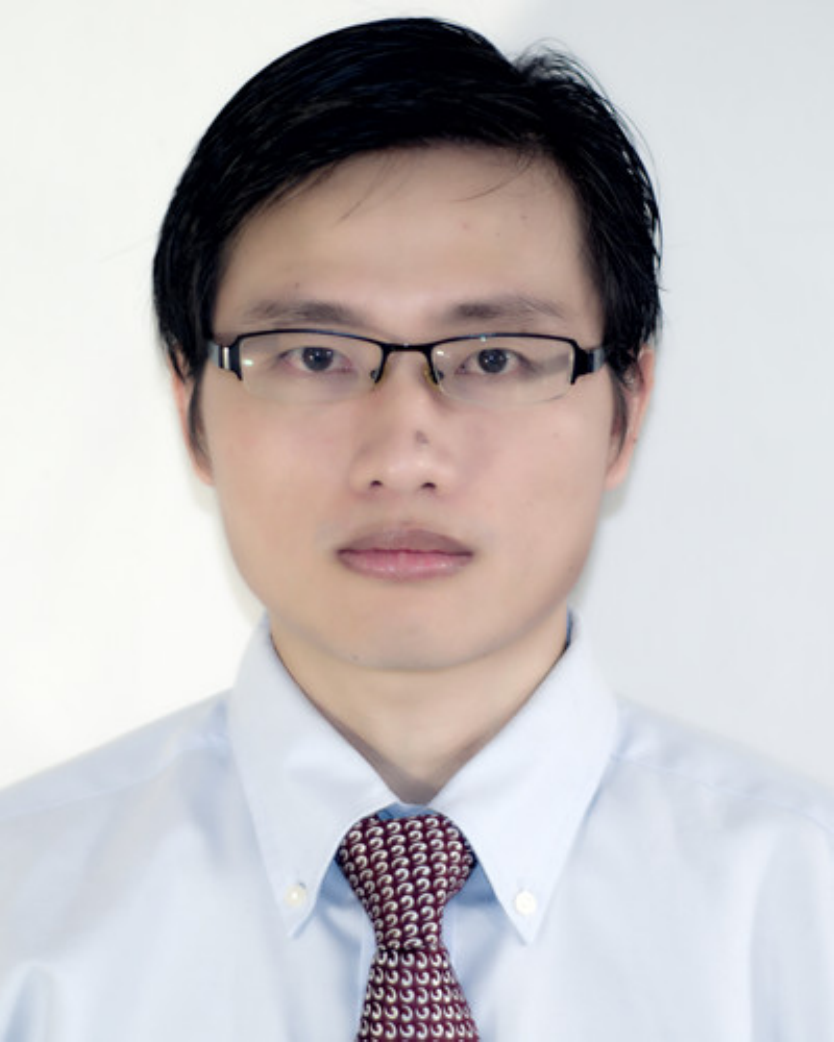}}]{Zhen-Peng Bian}
received the Bachelor degree from South China University of Technology, China, and the Ph.D. degree from Nanyang Technological University, Singapore, in 2007 and 2016, respectively. His research interests include video analytics for intelligent transportation system, video surveillance, human motion analysis, face detection and recognition, object detection, computer vision, and deep learning.
\end{IEEEbiography}

\begin{IEEEbiography}[{\includegraphics[width=1in,height=1.25in,clip,keepaspectratio]{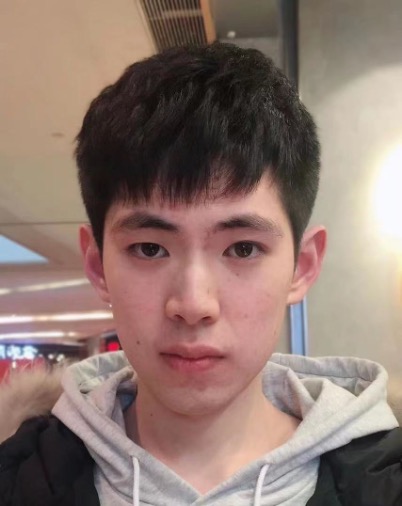}}]{Yunhao Zhou}
received the B.S. degree from the University of Electronic Science and Technology, Chengdu, China, in 2019. He is currently working toward the M.Eng. degree in the School of Electrical and Electronic Engineering, Nanyang Technological University, Singapore. His research interests include image retrieval and object tracking.
\end{IEEEbiography}

\begin{IEEEbiography}[{\includegraphics[width=1in,height=1.25in,clip,keepaspectratio]{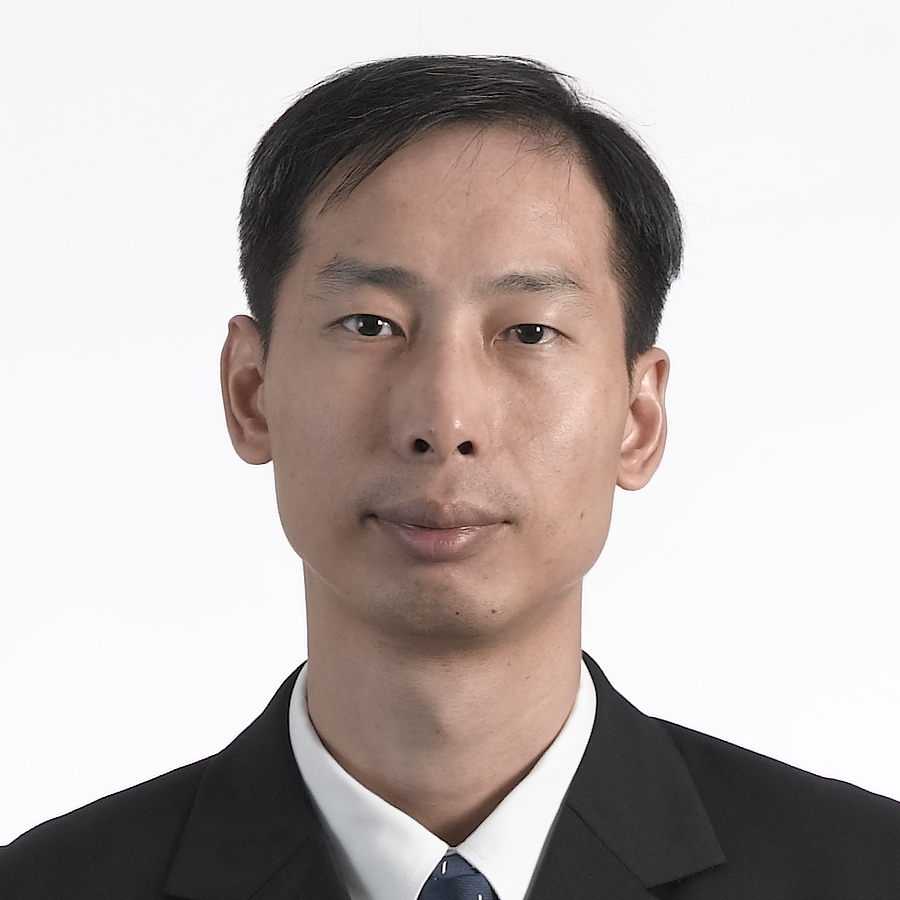}}]{Lap-Pui Chau}
received the Bachelor degree from Oxford Brookes University,  and the Ph.D. degree from The Hong Kong Polytechnic University, in 1992 and 1997, respectively.  He is Assistant Chair (Academic) of School of Electrical and Electronic Engineering, Nanyang Technological University. His research interests include fast visual signal processing algorithms, light-field imaging, video analytics for intelligent transportation system, and human motion analysis.

He was a General Chairs for IEEE International Conference on Digital Signal Processing (DSP 2015) and International Conference on Information, Communications and Signal Processing (ICICS 2015). He was a Program Chairs for International Conference on Multimedia and Expo (ICME 2016), Visual Communications and Image Processing (VCIP 2020, VCIP 2013), International Conference on Digital Signal Processing (DSP 2018) and International Symposium on Intelligent Signal Processing and Communications Systems (ISPACS 2010).

He was the chair of Technical Committee on Circuits \& Systems for Communications (TC-CASC) of IEEE Circuits and Systems Society from 2010 to 2012. He served as an associate editor for IEEE Transactions on Multimedia, IEEE Signal Processing Letters, IEEE Transactions on Circuits and Systems for Video Technology, IEEE Transactions on Circuits and Systems II. He is currently serving as an associate editor for IEEE Transactions on Broadcasting, and The Visual Computer (Springer Journal). Besides, he was an IEEE Distinguished Lecturer for 2009-2020, and a steering committee member of IEEE Transactions for Mobile Computing from 2011-2013. He is an IEEE Fellow.
\end{IEEEbiography}



\end{document}